\documentclass[10pt,twocolumn,letterpaper]{article}

\usepackage{wacv}            

\definecolor{wacvblue}{rgb}{0.21,0.49,0.74}
\usepackage[pagebackref,breaklinks,colorlinks,allcolors=wacvblue]{hyperref}

\def\wacvPaperID{456} 
\def\confName{WACV}
\def\confYear{2027}

\usepackage{amsmath,amsfonts}
\usepackage{amsopn}
\usepackage{bm} %
\usepackage{multirow}
\usepackage{tabularx}
\newcommand{\vct}[1]{\boldsymbol{#1}} %
\newcommand{\mat}[1]{\boldsymbol{#1}} %

\usepackage{booktabs}

\usepackage{xcolor}

\usepackage{xspace}

\newcommand{\OursP}{\method{Prompt-CAM}\xspace}
\newcommand{\OursI}{\method{INTR}\xspace}
\newcommand{\Ours}{INTR++\xspace}
\newcommand{\field}[1]{\mathbb{#1}}

\newcommand{\R}{\field{R}} %

\newcommand{\ProbOpr}[1]{\mathbb{#1}}

\newcommand{\expect}[2]{%
\ifthenelse{\equal{#2}{}}{\ProbOpr{E}_{#1}}
{\ifthenelse{\equal{#1}{}}{\ProbOpr{E}\left[#2\right]}{\ProbOpr{E}_{#1}\left[#2\right]}}} %
\newcommand{\var}[2]{%
\ifthenelse{\equal{#2}{}}{\ProbOpr{VAR}_{#1}}
{\ifthenelse{\equal{#1}{}}{\ProbOpr{VAR}\left[#2\right]}{\ProbOpr{VAR}_{#1}\left[#2\right]}}} %

\DeclareMathOperator{\argmax}{arg\,max}

\usepackage{cuted}

\newcommand{\vp}{\vct{p}}

\newcommand{\ve}{{\vct{e}}}
\newcommand{\vx}{{\vct{x}}}

\newcommand{\vz}{{\vct{z}}}

\newcommand{\vw}{\vct{w}}

\newcommand{\mP}{\mat{P}}

\newcommand{\mE}{\mat{E}}

\newcommand{\mV}{\mat{V}}
\newcommand{\mQ}{\mat{Q}}

\newcommand{\mZ}{\mat{Z}}

\newcommand{\mI}{\mat{I}}
\newcommand{\mK}{\mat{K}}

\newcommand{\method}[1]{\textsc{#1}}
\newcommand{\eat}[1]{}

\title{INTR++: INterpretable TRansformer with better Accuracy, Interpretability, and Adaptability}

\author{
Dipanjyoti Paul$^{1,3, 4}$\hspace{0.5em}
Arpita Chowdhury$^{2}$\hspace{0.5em}
Maneet Chatterjee$^{3}$\hspace{0.5em}
Daichi Kitaguchi$^{4}$\hspace{0.5em} \\
Tanya Berger-Wolf$^{2}$ \hspace{0.5em}
Tetsuya Sakurai$^{4}$\hspace{0.5em}
Wei-Lun Chao$^{5}$\\[0.2em]
{\small
$^{1}$\textbf{IIT Kharagpur }\qquad
$^{2}$\textbf{The Ohio State University} \qquad
$^{3}$\textbf{IIEST Shibpur }\qquad
$^{4}$\textbf{University of Tsukuba} \qquad
$^{5}$\textbf{Boston University}
}
}

\begin{document}

\twocolumn[{
\vspace{-11mm}
\maketitle
\vspace{-10mm}

\begin{center}
    \includegraphics[width=0.85\textwidth]{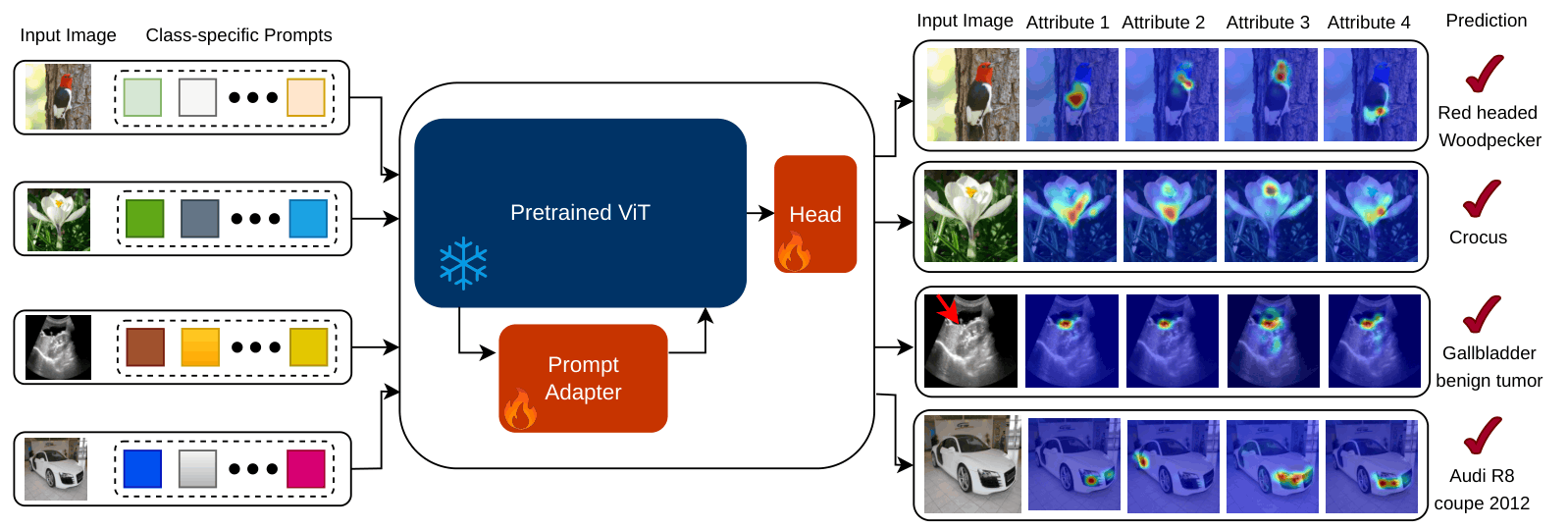}
\end{center}

\vspace{-3mm}

\noindent
\begin{minipage}{\textwidth}
\refstepcounter{figure}
\label{fig_teaser}

\small
\textbf{Figure \thefigure.} \textbf{Illustration of \Ours.}
It consists of learnable \textit{class-specific prompts }(one for each class) for identifying discriminative patterns, a frozen \textit{backbone} (ViT), and a learnable \textit{classifier head}. In addition, the learnable \textit{prompt adapter} module (PAM) enriches the prompts with domain-specific knowledge while making only minimal adjustments to the pretrained backbone representations. During training on each target dataset, PAM enables \Ours\ to capture discriminative patterns across input categories. This facilitates accurate predictions with supporting evidence across diverse domains, including those with substantial input variation such as medical imaging.
\end{minipage}

\vspace{4mm}
}]

\begin{abstract}\textit{Interpretable Transformers}  often perform well in domain specific settings but lose effectiveness beyond that. Achieving ViT-like broad applicability requires models with consistent performance across domains. Recent works on \textit{Interpretable Transformers}, INTR and Prompt-CAM, were developed keeping domain adaptability in mind. Prompt-CAM leverages learnable prompts to promote adaptability. However, these prompts are \textit{fixed} in nature  and therefore struggle to adapt to  domains with substantial variation in input image characteristics,  resulting in reduced interpretability. To address this limitation, we propose \textbf{\Ours}, which extends Prompt-CAM by incorporating a lightweight \textbf{Prompt Adapter} Module (PAM). PAM enables prompts to be \textit{adaptive} by enriching them with domain-specific knowledge, thereby ensuring consistent interpretability across diverse domains including those with varying image characteristics such as medical imaging, micro-scopic imaging, and remote sensing imagery. \Ours retains the standard architecture and end-to-end learning of \textit{Interpretable Transformer} based models, while adding only a small number of additional parameters compared to Prompt-CAM. We demonstrate that  \Ours delivers consistent improvements in all aspects across domains like animals, plants, medical imaging, microscopic imaging, remote sensing imagery and object-centric data.
\end{abstract}

\section{Introduction}

\textit{Interpretable Transformers}~\cite{paul2024simple, chowdhury2025prompt, mannix2024comfe} have recently gained significant attention due to their standard architectures and end-to-end learning strategies. These methods have demonstrated strong interpretability across diverse domains such as animals, plants, and object-centric datasets. We refer to these as domains with \textit{natural image} input. However, their adaptability to medical imaging, and microscopic imaging is limited. This limitation arises from the substantial variation in the image characteristics compared to natural images, and we refer to such domains as those with \textit{specialized image} input. Existing \textit{Interpretable Transformer} methods exhibit a decline in its adaptability 
when transitioning from domains with natural image input to those with specialized image input.

Beyond adaptability, another critical limitation lies in their classification performance. Although these methods demonstrate strong interpretability in domains with natural images, they often struggle to deliver high predictive accuracy. This shortcoming is commonly observed in prompt-based \textit{Interpretable Transformers}~\cite{chowdhury2025prompt}, where the class-specific prompts lack the capacity to capture fine-grained localized information essential for accurate prediction. This is due to the static nature of the prompts. By \textit{static}, we mean prompts that are learned solely through the attention mechanism of the frozen backbone. In contrast, we refer to prompts as \textit{adaptive} when they are learned not only through the  backbone but also through an adapter module, enriching them with domain-specific knowledge.

This highlights the need for \textit{Interpretable Transformers} that effectively handle the trade-off between classification performance and interpretability irrespective of the input image type. Limitations in existing models stem from the frozen backbone, which restricts the prompts from becoming truly adaptive. Nevertheless, these backbones still encode useful spatial structures that, if leveraged appropriately, can be beneficial. Therefore, building upon existing prompt-based Interpretable Transformers, we design a new framework that addresses two key concerns:
$1$) enabling prompting strategies in prompt-based Interpretable Transformers to adapt to both natural and specialized image inputs while keeping the pretrained backbone intact; and
$2$) providing meaningful interpretations while simultaneously achieving high classification accuracy.

Recent advances in Vision Transformers (ViTs)~\cite{dosovitskiy2021an} have significantly improved visual recognition, particularly for fine-grained analysis~\cite{wang2023open, tang2023weakly, zhu2022dual, he2022transfg}. Self-supervised models such as DINO~\cite{caron2021emerging}, DINOv2~\cite{oquab2023dinov2}, and DINOv3~\cite{simeoni2025dinov3}, trained on large-scale natural image datasets, exhibit strong object-level localization capabilities. Therefore, these self-supervised models constitute strong backbone candidates, as they provide semantically meaningful representations that are well suited for interpretable Transformers.

To this end, we propose \Ours, which aims to perform well across domains irrespective of input type while simultaneously improving both accuracy and interpretability. \Ours builds upon Prompt-CAM by introducing a lightweight adapter module that enhances adaptability and classification performance without compromising interpretability. The introduced Prompt Adapter Module (PAM) is strategically integrated to enrich the prompts with domain-specific knowledge while making only minimal adjustments to the pretrained backbone representations. 
Other adapter approaches such as LoRA~\cite{hu2022lora} and AdaptFormer~\cite{chen2022adaptformer} introduce adapter components for domain generalization; however, their designs limit the ability to align prompt representations with salient input characteristics in interpretability-driven settings.
Pls. refer to \autoref{embeddings} for more details.

\begin{figure*}[!t]
  \centering
   \includegraphics[width=0.88\linewidth]{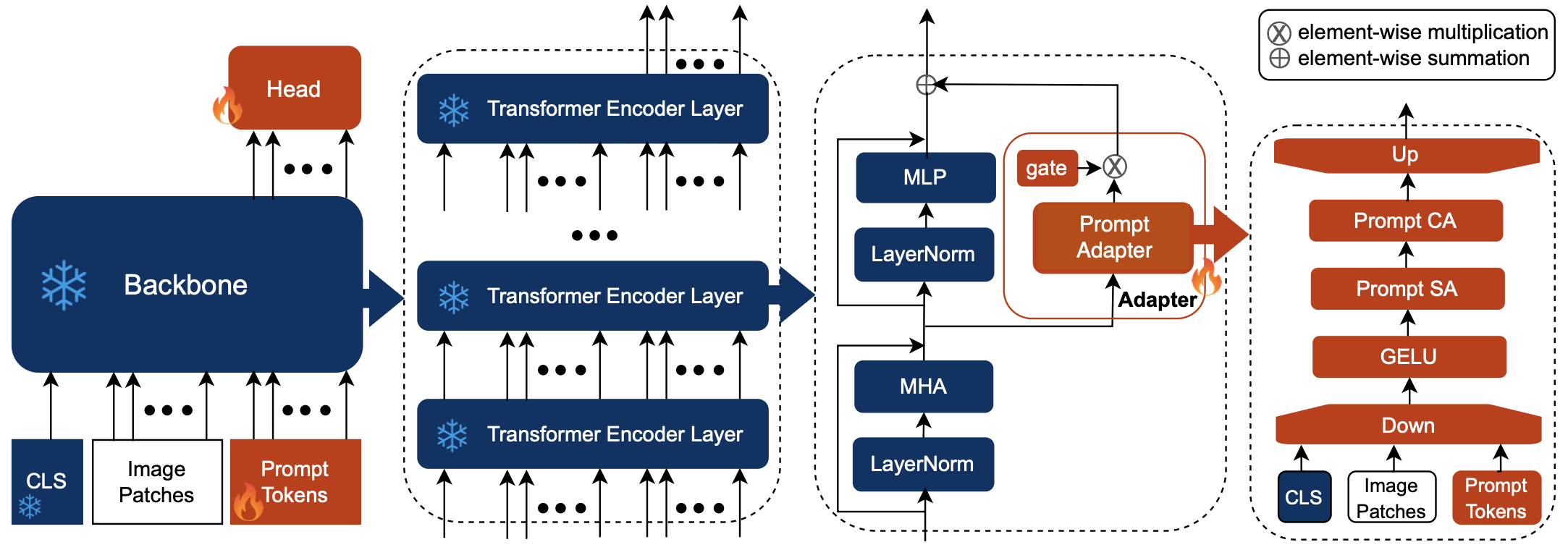}
   \vskip -3mm
   \caption{\textbf{Architecture of \Ours.} \Ours employs a pretrained frozen backbone, while learning only the prompts and a lightweight prompt adapter module. The module is simple, adding only a small number of additional parameters. It consists of a lower-dimensional linear projection, a non-linear activation, self-attention over prompt tokens, cross-attention with image patches, and a projection back to the original dimension. The adapted prompt tokens at the last layer are finally used for classification. See the Method section for details.}
   \label{flowchart}
   \vskip -2mm
\end{figure*}

The \textit{Prompt Adapter} integrated into \Ours enriches the prompt representations by incorporating domain-specific knowledge, while introducing only minimal residual adjustments to the backbone patch representations, thereby preserving the pretrained visual representations. The details of the introduced adapter module are illustrated in ~\autoref{flowchart}. The module is inserted into each transformer layer and operates on the intermediate token representations. A linear transformation followed by a non-linear activation enables the extraction of domain-specific features. The cross-attention mechanism between prompt tokens and image patches facilitates the integration of salient domain-specific characteristics into the prompts, thereby enabling \Ours to deliver accurate predictions alongside meaningful interpretability across diverse domains.

\Ours addresses the challenges in existing prompt-based \textit{Interpretable Transformers} by incorporating a lightweight module that enhances  accuracy, interpretability and adaptability while keeping the model computationally efficient. It also addresses the trade-off between accuracy and interpretability. We extensively evaluate \Ours on ten  datasets across a diverse range of domains:
\textbf{Animal:}~\cite{wah2011caltech, khosla2011novel};
\textbf{Plant:}~\cite{nilsback2008automated, roopashree2020medicinal};
\textbf{Object-centric:}~\cite{krause20133d, maji2013fine};
\textbf{Medical imaging:}~\cite{nickparvar2024, basu2022surpassing};
\textbf{Microscopic imaging:}~\cite{zhuhard, komura2020histology};
\textbf{Remote sensing imagery:}~\cite{xia2017aid, di2021public}. Our contributions are two-fold.

\begin{itemize}
\item We introduce a lightweight Prompt Adapter Module (PAM), that enhances accuracy and interpretability while enabling effective domain adaptability across both natural and specialized images.
\item We demonstrate the effectiveness of \Ours through extensive experiments on datasets spanning domains with both natural and specialized image characteristics. \Ours also effectively handles the trade-off between accuracy and interpretability observed in existing prompt-based approaches. The source code is available at \noindent \url{https://github.com/dipanjyoti/INTR-PlusPlus}.
\end{itemize}
    
\section{Related Work}
\label{sec:formatting}

Meaningful interpretation is crucial for fine-grained tasks in domains such as biodiversity and healthcare. Early works focused on post-hoc explainability, where predictions are retrospectively analyzed to identify influential regions. Grad-CAM~\cite{selvaraju2017grad}, Layer-CAM~\cite{jiang2021layercam}, Eigen-CAM~\cite{muhammad2020eigen}, Attention Roll-out~\cite{kashefi2023explainability}, GradCAM++~\cite{Chattopadhay_2018}, Ablation-CAM~\cite{ramaswamy2020ablation}, and XGradCAM~\cite{fu2020axiombasedgradcamaccuratevisualization} generate saliency maps using gradients. However, these methods often lack trustworthiness as they explain decisions only after the prediction is made.

In contrast, interpretable methods identify key attributes during the prediction process itself. Methods such as ProtoPFormer~\cite{xue2022protopformer}, TesNet~\cite{wang2021interpretable}, ProtoConcepts~\cite{ma2024looks}, ProtoViT~\cite{ma2024interpretable}, and PIPNet~\cite{nauta2023pip} use prototype-based reasoning to generate interpretations. However, they typically require complex architectures and multi-stage training, which limit their adaptability, applicability, and reproducibility.

To address these challenges, a new line of work, namely \textit{Interpretable Transformers}, including \OursI~\cite{paul2024simple} and \OursP~\cite{chowdhury2025prompt}, has been proposed. The core idea is to provide meaningful interpretation using standard architectures and end-to-end learning objectives. However, these methods are tightly coupled with the backbone and struggle to generalize to specialized image inputs. Moreover, these models suffer from a trade-off between accuracy and interpretability. Additionally, the prompts in Prompt-CAM are \textit{static} in nature, which limits their adaptability when applied to domains beyond natural images. Therefore, there is a need for \textit{adaptive} prompts.

Recent works on domain adaptation, such as AdaptFormer~\cite{chen2022adaptformer}, LoRA~\cite{hu2022lora}, and DVPT~\cite{he2025dvpt}, attempt to address cross-domain challenges. However, the designs of AdaptFormer and LoRA are not suitable for refining prompt representations to better capture salient input characteristics. In contrast, DVPT relies on the [CLS] token, which limits its ability to capture class-specific attributes.

\section{Method}

In this paper, we introduce \textbf{\Ours}, an \textit{Interpretable Transformer} framework inspired by Prompt-CAM~\cite{chowdhury2025prompt} and Visual Prompt Tuning (VPT)~\cite{jia2022visual}, which incorporates a lightweight Prompt Adapter Module (PAM). The overall architecture remains simple, as the adapter can be easily integrated into transformer layers while fully leveraging the backbone model. The architecture is illustrated in \autoref{flowchart}. For clarity, we follow the same naming conventions and notation as Prompt-CAM.

\subsection{Vision Transformer Preliminaries}
Vision Transformer (ViT)~\cite{vaswani2017attention} serves as the backbone of our approach. An input image $\mI$ is divided into $M$ fixed-size patches, each linearly projected into a $D$-dimensional embedding. Positional embeddings are added to form the input sequence $\mE_0 = [\ve_0^1, \cdots, \ve_0^M] \in \R^{D \times M}$. A learnable [CLS] token $\vx_0 \in \R^D$ is prepended, resulting in $\mE = [\vx_0, \mE_0]$. The transformer comprises $N$ layers, each consisting of a Multi-Head Attention (MHA) block, a Multi-Layer Perceptron (MLP), and Layer Normalization (LN). Traditionally, the final [CLS] token, $\vx_N$, is used for classification. The MHA operation is defined as
\begin{align}
\mathrm{MHA}(q,k,v) 
   = \mathrm{Softmax}\!\left(\frac{\mQ  \mK^\top }{\sqrt{D}}\right)\mV , \label{softmax_ViT}
\end{align}
where $q$, $k$, and $v$ denote the query, key, and value vectors derived from the input token embeddings $[\vx, \mE]$, and $\mQ$, $\mK$, and $\mV$ represent their linear projections using the corresponding weights $W_q$, $W_k$, and $W_v$, respectively:
\begin{align}
\mQ = W_q q, \quad \mK = W_k k, \quad \mV = W_v v. \label{eq_proj}
\end{align}

\subsection{Overall Workings of \Ours}
\Ours is built  on the deep variants of \OursP~\cite{chowdhury2025prompt}. It leverages pretrained ViT backbones for classification. The PAM is integrated into each layer of the backbone model. While the backbone is kept frozen, PAM remains learnable. Additionally, for a dataset with $C$ classes, \Ours introduces $C$ learnable class-specific prompt vectors, each of dimension $D$. The prompts are injected at each layer and learned independently. Prompts from the final layer are used as class-specific tokens for prediction.

Given a labeled training example $(\mI, y \in \{1,\cdots, C\})$, unlike traditional ViT classifiers~\cite{dosovitskiy2020image, liu2021swin}, \Ours relies on class-specific prompts injected at the last layer, denoted as $\mP_{N-1} = [\vp^1, \cdots, \vp^C]$, and produces class-specific outputs $\mZ_N = [\vz_N^1, \cdots, \vz_N^C]$. The logits are computed as:
\begin{align}
s[c] = \vw^\top \vz_N^c, \quad 1 \leq c \leq C,
\label{eqn_logits1}
\end{align}
where $\vw \in \R^D$ is a shared learnable vector. This formulation makes \Ours naturally suited for interpretation, as it enables the analysis of attention maps between prompts and image patches. The class-specific prompts $\mP$ are trained by minimizing the standard cross-entropy loss. Finally, the predicted label $\hat{y}$ is obtained as:
\begin{align}
\hat{y} = \argmax_{c \in \{1,\cdots, C\}} s[c]. \notag
\end{align}

\subsection{Prompt Adapter Module (PAM)}
\label{ss_domain_aware_module}
We describe the workings of transformer layers in \Ours, with a focus on the  \textit{PAM}, an important aspect of \Ours. At the $i$-th layer, the model takes as input the learnable prompt tokens $\mP_{i-1}$, patch embeddings $\mE_{i-1}$, and the [CLS] token $\vx_{i-1}$. These are concatenated and passed through a standard Multi-Head Attention (MHA) block.
\begin{align}
[\mZ_i', \mE_i', \vx_i'] &= \mathrm{MHA}\!\left(\mathrm{LN}([\mP_{i-1}, \mE_{i-1}, \vx_{i-1}])\right)  \label{mha}   \\
&\quad + [\mP_{i-1}, \mE_{i-1}, \vx_{i-1}], \quad i = 1, \cdots, N. \notag 
\end{align}

Here, $\mZ_i^{'}$ denotes the intermediate prompt output. The Prompt Adapter Module (PAM), operates on the MHA output, taking $[\mZ_i^{'}, \mE_i^{'}, \vx_i^{'}]$ as input as in~\autoref{mha}. The PAM, first projects all input tokens to a lower-dimensional space using a compression layer $f_{dn}$ as
\begin{align}
[\hat{\mZ}_i^{d}, \hat{\mE}_i^{d}, \hat{\vx}_i^{d}] &= \sigma(f_{dn}([\mZ_i^{'}, \mE_i^{'}, \vx_i^{'}])), \  i = 1, \cdots, N.   \notag
\end{align}
Here, $f_{dn}$ is a learnable linear layer with weights $\vw \in \R^{D \times d}$, where $d \ll D$ and $\sigma$ is a \textit{GELU} activation function. 
We  split the tokens into prompt, patch and [CLS] as
\begin{align}
\hat{\mZ}_i^{d}, [\hat{\mE}_i^{d}, \hat{\vx}_i^{d}] &= \mathrm{split}([\hat{\mZ}_i^{d}, \hat{\mE}_i^{d}, \hat{\vx}_i^{d}]), \ \ \ i = 1, \cdots, N. \notag
\end{align}
Next, we apply Multi-Head Attention (MHA) referring to~\autoref{softmax_ViT},  within PAM. 
First, we apply Prompt Self-Attention (PSA) to the prompt tokens $\hat{\mZ}_i^d$ as 
\begin{align}
\bar{\mZ}_i^d = \mathrm{MHA}(q= k= v=\hat{\mZ}_i^d).
\label{psa}
\end{align}
PSA allows prompts to share information among themselves, which is useful when classes have overlapping features. To  enrich prompts  to capture better salient input characteristics, we  perform Prompt Cross-Attention (PCA) using patch and [CLS] tokens as keys and values
\begin{align}
\textcolor{red}{\Tilde{\mZ}_i^d} 
   &= \mathrm{MHA}(q=\bar{\mZ}_i^d, k= v=[\hat{\mE}_i^d, \hat{\vx}_i^d]).
   \label{pca}
\end{align}
The prompts $\textcolor{red}{\Tilde{\mZ}_i^d}$  are now enriched with the domain specific knowledge. 
We then concatenate the updated prompts and patch tokens as $[\textcolor{red}{\Tilde{\mZ}_i^d}, \hat{\mE}_i^d, \hat{\vx}_i^d]$ and project them back to the original dimension $D$ using a learnable linear layer $f_{up}$ with weights $\vw \in \R^{d \times D}$, modulated by a gating parameter $g$ as
\begin{align}
[\textcolor{red}{\mZ_i^{p}},\; 
 \textcolor{blue}{\mE_i^{p}},\; 
 \textcolor{blue}{\vx_i^{p}}] = g \cdot f_{up}([\textcolor{red}{\Tilde{\mZ}_i^d}, \hat{\mE}_i^d, \hat{\vx}_i^d]),  i = 1,... , N,
\end{align}
here, \textit{gate}, $g$ is a learnable scalar parameter that controls the flow of information from the adapter, and the superscript $(\cdot)^p$ denotes PAM adapter. These are then fused with the transformer's intermediate features to produce the layer's output as
\begin{align}
[\mZ_i, \mE_i, \vx_i] &= \mathrm{MLP}(\mathrm{LN}([\mZ_i^{'}, \mE_i^{'}, \vx_i^{'}])) \notag \\
&\quad + [\mZ_i^{'}, \mE_i^{'}, \vx_i^{'}] 
\label{eqn_fusion_PAM} \\
&\quad + [\textcolor{red}{\mZ_i^{p}},\; 
 \textcolor{blue}{\mE_i^{p}},\; 
 \textcolor{blue}{\vx_i^{p}}], \quad i = 1, \cdots, N. \notag
\end{align}

The red \& blue highlighted terms in~\autoref{eqn_fusion_PAM} capture the adaptive behavior introduced by PAM. Specifically, $\textcolor{blue}{[\mE_i^p, \vx_i^p]}$ introduces only minimal residual adjustments to the patch  token representations, while $\textcolor{red}{\mZ_i^p}$ enriches the prompt representations with domain-specific knowledge. Together, these enable \Ours to generalize effectively across both natural and specialized image domains.

\subsection{Prompt Adapter  vs. Others}

We evaluate the proposed adapter technique against several alternative adaptation strategies, including a  MLP-based adapter and two widely used  adaptation methods, AdaptFormer~\cite{chen2022adaptformer} and LoRA~\cite{hu2022lora}. For a fair comparison, each adaptation module is consistently applied to both the input image tokens and the input prompts. See Suppl. Sec. \ref{adaption_Suppl}.


Like PAM, AdaptFormer operates after the MHA block  and takes $[\mZ_i', \mE_i', \vx_i']$ as input (see~\autoref{mha}).   However, unlike PAM, AdaptFormer treats all tokens uniformly, without giving prompts any special emphasis.   It performs a down-projection, followed by a ReLU activation, and then an up-projection back to the original dimension as
\begin{align}
\textcolor{blue}{[\mZ_i^{a}, \mE_i^{a}, \vx_i^{a}]}
   &= g \cdot f_{up}\!\left(\sigma\!\left(f_{dn}([\mZ_i^{'}, \mE_i^{'}, \vx_i^{'}])\right)\right), \\
   &\quad i = 1, \cdots, N. \notag
\end{align}

The superscript $(\cdot)^a$ denotes AdaptFormer outputs. 
The resulting \textcolor{blue}{$[\mZ_i^{a}, \mE_i^{a}, \vx_i^{a}]$} incorporates domain-specific cues, 
but since AdaptFormer does not explicitly adapt prompts, its effectiveness in prompt-based classification remains limited. 
It is then passed through the remaining transformer layer, producing the final output as in~\autoref{eqn_fusion_PAM}, 
with \textcolor{blue}{$[\mZ_i^{a}, \mE_i^{a}, \vx_i^{a}]$} 
 as the adapter contribution.

In LoRA, the adaptation operates differently. It introduces low-rank adapters within the MHA block itself.  Thus, we rewrite~\autoref{mha} for LoRA as
\begin{align}
\textcolor{blue}{[\mZ_i^{l}, \mE_i^{l}, \vx_i^{l}]} 
   &= \mathrm{MHA}^{\text{LoRA}}( \mathrm{LN}([\mP_{i-1}, \mE_{i-1}, \vx_{i-1}]) ) \label{mha88} \\
   &\quad + [\mP_{i-1}, \mE_{i-1}, \vx_{i-1}], \quad (i=1,\ldots,N). \notag 
\end{align}

Here, the superscript $(\cdot)^l$ in \textcolor{blue}{$[\mZ_i^{l}, \mE_i^{l}, \vx_i^{l}]$}, denotes the LoRA outputs, which encode domain-specific information through LoRA multi-head attention ($\mathrm{MHA}^{\text{LoRA}}$) as
\begin{align}
\mathrm{MHA}^{\text{LoRA}}(q,k,v) 
   &= \mathrm{Softmax}\! \left(\frac{Q^{l}(K^{l})^\top }{\sqrt{D}}\right) V^{l}, \label{softmax_LoRA}
\end{align}
with the projections in~\autoref{eq_proj} modified as
\begin{align}
Q^{l} &= (W_q + \Delta W_q)\, q, \quad \Delta W_q = A_q B_q, \notag \\
K^{l} &= (W_k + \Delta W_k)\, k, \quad \Delta W_k = A_k B_k, \\
V^{l} &= (W_v + \Delta W_v)\, v, \quad \Delta W_v = A_v B_v, \notag
\end{align}

where $A_q \in \mathbb{R}^{D \times d}$ and $B_q \in \mathbb{R}^{d \times D}$ 
with $d \ll D$; similar definitions hold for $A_k, B_k$ and $A_v, B_v$.  
The final output of the transformer layer is:
\begin{align}
[\mZ_i, \mE_i, \vx_i] &= \mathrm{MLP}(\mathrm{LN}(\textcolor{blue}{[\mZ_i^{l}, \mE_i^{l}, \vx_i^{l}]})) \notag \\
&\quad + \textcolor{blue}{[\mZ_i^{l}, \mE_i^{l}, \vx_i^{l}]}. 
\label{eqn_fusion}
\end{align}

In LoRA, the original weights $W_q, W_k, W_v$ are frozen, and only the low-rank adapters 
$A$ and $B$ (i.e., $\Delta W$) are learnable. Thus, LoRA injects domain-specific information.


In summary, the outputs of all adaptive modules $[\textcolor{red}{\mZ_i^{p}}, \textcolor{blue}{\mE_i^{p}}, \textcolor{blue}{\vx_i^{p}}]$,
\textcolor{blue}{$[\mZ_i^{a}, \mE_i^{a}, \vx_i^{a}]$},
and \textcolor{blue}{$[\mZ_i^{l}, \mE_i^{l}, \vx_i^{l}]$} enrich the prompt representations with domain-specific knowledge while introducing only minimal residual adjustments to the backbone representations.
This behavior is also observed when using a simple MLP as the adapter. However, PAM differs fundamentally in how this adaptation is structured and utilized.  This design of PAM enables prompts to not only absorb domain-specific knowledge implicitly, but also to actively align themselves with salient input characteristics. In contrast, AdaptFormer and LoRA enrich prompts indirectly through shared token adaptation, without explicitly modeling prompt \& patch interactions or enforcing prompt-specific enrichment. For all adaptive modules, the intermediate dimensionality is fixed at $d=64$, which was empirically found to provide a favorable trade-off between capacity and efficiency. More in the Suppl. Sec.~\ref{projection_dimension}.

\begin{figure}[!htbp]
  \centering
   \includegraphics[width=0.91\linewidth]{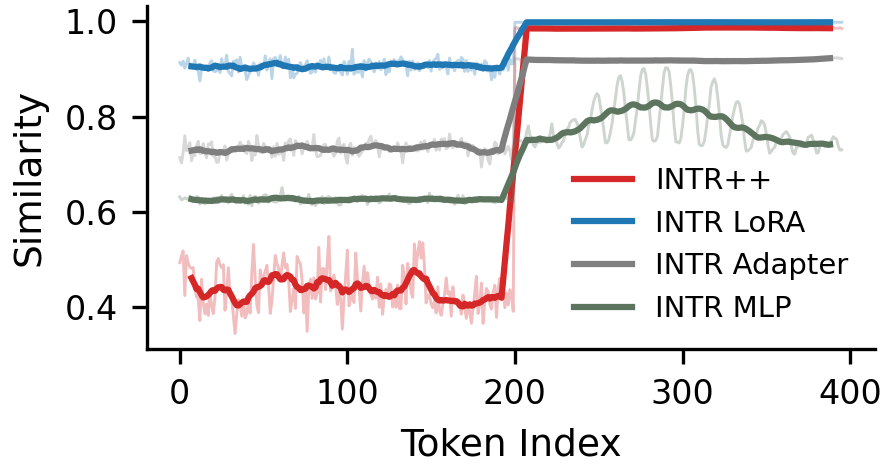}
   \vskip -4mm
   \caption{\textbf{Information induced by different adapters.} The first $200$ tokens in the plot are learnable prompt tokens for CUB dataset, while the remaining tokens are image patches. The bold curves represent the exponential moving average (EMA) of the raw cosine similarity values computed using a span of fifteen.}
   \label{embeddings}
   \vskip -4mm
\end{figure}

\autoref{embeddings} illustrates the cosine similarity between token representations before and after adapter insertion. As shown, the MLP substantially disrupts the backbone representations, making it unsuitable for prompt adaptation. In contrast, LoRA, PromptAdapter, and \Ours introduce only minimal adjustments to the backbone image patch representations, thereby preserving the pretrained visual representations. Notably, \Ours yields significantly richer prompt representations than LoRA and AdaptFormer, suggesting more effective incorporation of domain-specific knowledge. 
This behavior indicates that the Prompt Adapter Module (PAM) acts as a controlled residual correction mechanism, selectively enriching the prompt representations without substantially perturbing the pretrained backbone representations, making \Ours well suited for adaptation across both natural and specialized image domains.

\begin{figure}[!htbp]
  \centering
   \includegraphics[width=0.7\linewidth]{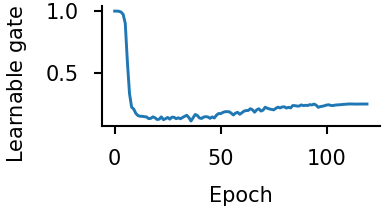}
   \vskip -4mm
   \caption{\textbf{Learned gate values over training epochs.} The plot shows the last layer gate values of \Ours on CUB dataset.}
   \label{gate}
   \vskip -2mm
\end{figure}

\subsection{Importance of the \textit{gate} Value} 


The learnable scaler gate $g$ in the PAM controls the module's contribution. It is initialized to one to promote rapid prompt adaptation to the target domain during early training. As shown in~\autoref{gate}, the model initially relies on stronger adaptive corrections and gradually reduces the adapter's contribution as the representations become aligned. More in the Suppl. Sec.~\ref{gate_value}.

\subsection{Feature Identification and Interpretation}
As shown in~\autoref{eqn_logits1}, the model predicts the score for class $c$ using the class-specific prompt output $\vz_N^{c}$. Since the projection vector $\vw$ is shared across all prompts, it does not introduce any class-specific bias. This design encourages each prompt to encode distinctive features relevant to its class, making them inherently discriminative.

By analyzing $\vz_N^{c}$, we can thus identify which features the model relies on for predicting class $c$. To provide interpretability, we examine the multi-head attention weights in the final transformer layer. Specifically, we extract the attention maps between each class-specific prompt and the image patch tokens. These maps are then superimposed on the input image to highlight the regions most influential to the model’s prediction. This serves as supporting visual evidence for model interpretability. Like \OursP, we greedily select top attention heads that are sufficient to distinguish between different categories, and  visualize only those in the experiment section.

\subsection{Additional Parameters Introduced by  PAM} Our adaptive module PAM is lightweight, adding only a small number of parameters. The two projection layers contribute $2(D \times d) + D + d$ parameters, where $d \ll D$ (e.g., $d=64$ and $D=768$ for ViT-Base). Parameterized multi-head attention (MHA) adds $4d^2$ for self-attention and $4d^2$ for cross-attention, totaling $8d^2$. Thus, the total overhead is $2(D \times d) + D + d + 8d^2$. For example, for the CUB~\cite{wah2011caltech} dataset with $200$ categories, the number of learnable parameters for \OursP is $\approx 2\%$ of the total parameters, while \Ours has only $\approx 3\%$. More in the Suppl. Sec.~\ref{computation}.

\section{Experiments}
\subsection{Experimental Setup}
\textbf{Dataset.}
We extensively evaluate the performance and adaptability of \Ours across diverse datasets spanning both \textit{natural image domains} and \textit{specialized image domains}. The natural image domains include \textbf{Animal:} CUB-200-2011 (\textit{CUB})~\cite{wah2011caltech} and Stanford Dogs (\textit{Dog})~\cite{khosla2011novel}; \textbf{Plant:} Oxford Flowers (\textit{Flower})~\cite{nilsback2008automated} and Medicinal Leaf (\textit{MedLeaf})~\cite{roopashree2020medicinal}; and \textbf{Object-centric:} Stanford Cars (\textit{Car})~\cite{krause20133d} and FGVC Aircraft (\textit{Craft})~\cite{maji2013fine}. The specialized image domains include \textbf{Medical:} MRI Brain Tumors (\textit{Brain})~\cite{nickparvar2024} and Gallbladder Cancer Ultrasound (\textit{GBCU})~\cite{basu2022surpassing}; \textbf{Microscopic:} Colon Cancer Whole Slide Imaging (\textit{Colon})~\cite{zhuhard}; and \textbf{Remote Sensing:} Aerial Image Dataset (\textit{AID})~\cite{xia2017aid}. More in the Suppl. Sec.~\ref{dataset_suppl}.


\textbf{Model.} We utilize different backbones for \Ours, including DINO~\cite{caron2021emerging}, DINOv2~\cite{oquab2023dinov2}, and BioCLIP2~\cite{gu2025bioclip}. Unless stated otherwise, all results are based on the \textit{ViT-B} variant of DINO. The backbone is kept frozen in \Ours, while only the prompts, PAM, and the classification head are learned. More in the Suppl. Sec.~\ref{model_suppl}.

\textbf{Baselines.} We compare \Ours with several baseline methods. We show comparison with  the interpretable methods like  TesNet~\cite{wang2021interpretable}, ProtoConcepts~\cite{ma2024looks}, PIPNet~\cite{nauta2023pip}, ProtoViT~\cite{ma2024interpretable}
 and \OursP~\cite{chowdhury2025prompt}. Additionally, we compare with post-hoc methods such as Grad-CAM~\cite{selvaraju2017grad}, Layer-CAM~\cite{jiang2021layercam},  Eigen-CAM~\cite{muhammad2020eigen}, and Attention Roll-out~\cite{kashefi2023explainability}.
 Finally, to evaluate the contribution of the integrated adaptive module, we compare \Ours with adaptation based approaches like LoRA~\cite{hu2022lora} and AdaptFormer~\cite{chen2022adaptformer}.

\subsection{Overall Performance of \Ours}

\Ours has been extensively evaluated across datasets from diverse domains, demonstrating its effectiveness in three key aspects: improved accuracy, better interpretability and adaptability. As shown in~\autoref{fig_teaser}, the model consistently provides meaningful interpretations regardless of the domain. 
For biodiversity tasks, it identifies discriminative traits, while in medical imaging, it highlights relevant regions of interest. 
Furthermore, the improved classification performance of \Ours is evidenced by its gains over Prompt-CAM in~\autoref{tab_overall_acc}, while its superior interpretability is illustrated in~\autoref{fig_Prompt_INTRAI}. More in the Suppl. Sec.~\ref{accu_suppl}.

\subsection{Performance on  Specialized Image Domains}
The GBCU dataset is used as the primary reference to demonstrate the effectiveness of our approach, while similar trends hold across the other specialized image datasets.

\textbf{Expert evaluation.} We conducted a human evaluation on the heatmaps generated by \Ours. For the GBCU dataset, we randomly selected 10 images from the malignant and benign categories and collected corresponding heatmaps from \Ours, \OursP, Attention Rollout, and LayerCAM. These heatmaps were shared with five medical experts without disclosing the method used. The results show that \Ours accurately localized the region of interest in 91\% of the cases, compared to 73\% for \OursP, 69\% for Attention Rollout, and 47\% for LayerCAM. This highlights the effectiveness of \Ours. 

\begin{table}[!htbp]
\vskip-5pt
\centering
\footnotesize
\begin{tabular}{lcccc}
\toprule
\multirow{2}{*}{Method} & \multicolumn{2}{c}{CUB} & \multicolumn{2}{c}{GBCU} \\
                        & Insertion$\uparrow$ & Deletion$\downarrow$ & Insertion$\uparrow$ & Deletion$\downarrow$ \\
\midrule
Grad-CAM          & 0.52 & 0.17 & 0.57 & 0.26 \\
Layer-CAM         & 0.54 & 0.13 & 0.44 & 0.37 \\
Eigen-CAM         & 0.56 & 0.22 & 0.45 & 0.39 \\
Attention Rollout & 0.55 & 0.27 & 0.51 & 0.26 \\
\OursP                               & 0.61 & 0.09 & 0.59 & 0.21 \\
\textbf{\Ours}                               & \textbf{0.62} & \textbf{0.08} & \textbf{0.65} & \textbf{0.19} \\
\bottomrule
\end{tabular}
\caption{\small \textbf{Quantitative evaluation of interpretation. } Lower deletion and  higher  insertion scores indicate  better performance. }
\vskip-5pt
\label{tab_insertion_deletion}
\end{table}

\textbf{Quantitative interpretability evaluation.} To quantitatively evaluate the interpretation capability of \Ours, we use standard post-hoc interpretability metrics: insertion and deletion~\cite{petsiuk1806rise}. We compare \Ours with other interpretable as well as post-hoc methods. The~\autoref{tab_insertion_deletion} reports the scores on the GBCU medical imaging dataset, where \Ours consistently outperforms others in identifying important regions. This superiority is not limited to medical imaging; results on the CUB dataset further demonstrate its effectiveness on  domains with natural input image types.

\begin{table*}[!htbp]
  \centering
  {\small{
  \begin{tabular}{@{}lccccccccccc@{}}
    \toprule
    Method & CUB  & Dog & MedLeaf & Flower & GBCU & Brain & Colon & Car & Craft & AID\\
    \midrule
    Prompt-CAM & 73.2  & 81.1 & 96.9 & 86.8 & 63.1 & 93.5 & 74.2 & 83.2 & 68.7 & 93.5\\
    \Ours & \textbf{79.0} &  \textbf{83.5} & \textbf{97.1}& \textbf{87.2} &  \textbf{81.2} & \textbf{99.2} & \textbf{84.4} & \textbf{88.1} & \textbf{75.4} & \textbf{97.7}\\
    \bottomrule
  \end{tabular}
  }}
  \vskip -2mm
  \caption{\textbf{Accuracy comparison.} Results are on the DINO backbone. \Ours outperforms \OursP across domains.}
  \label{tab_overall_acc}
\end{table*}



\begin{figure}[!htbp]
  \centering
   \includegraphics[width=0.96\linewidth]{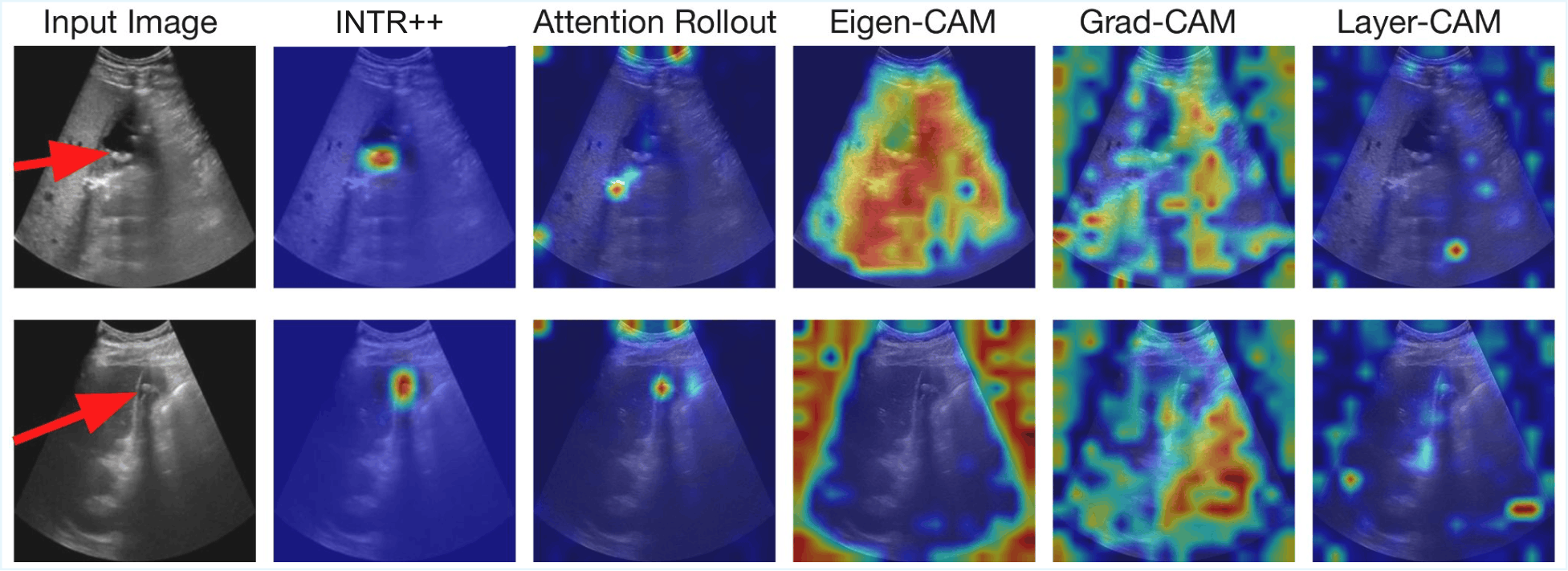}
   \vskip -2mm
   \caption{\textbf{Qualitative interpretability.} Comparison between \Ours and other post-hoc methods on the GBCU dataset.}
   \vskip -2mm
   \label{comparision_posthoc}
\end{figure}

\textbf{Qualitative interpretability analysis}
We qualitatively compare the interpretability of \Ours
 with post-hoc methods such as Attention Rollout, Eigen-CAM, Grad-CAM, and Layer-CAM. For \Ours, we use the top-ranked attention head, which alone is sufficient for correct prediction. ~\autoref{comparision_posthoc} illustrates examples from the benign category of the GBCU dataset, highlighting the relevant regions. \Ours clearly outperforms the other methods. More in the Suppl. Sec.~\ref{posthoc_suppl},~\ref{com_Sp_interp} and ~\ref{Interpretability_Comparisons}.

\subsection{Performance on  Natural Image Domains}

\begin{figure}[!htbp]
  \centering
   \includegraphics[width=0.98\linewidth]{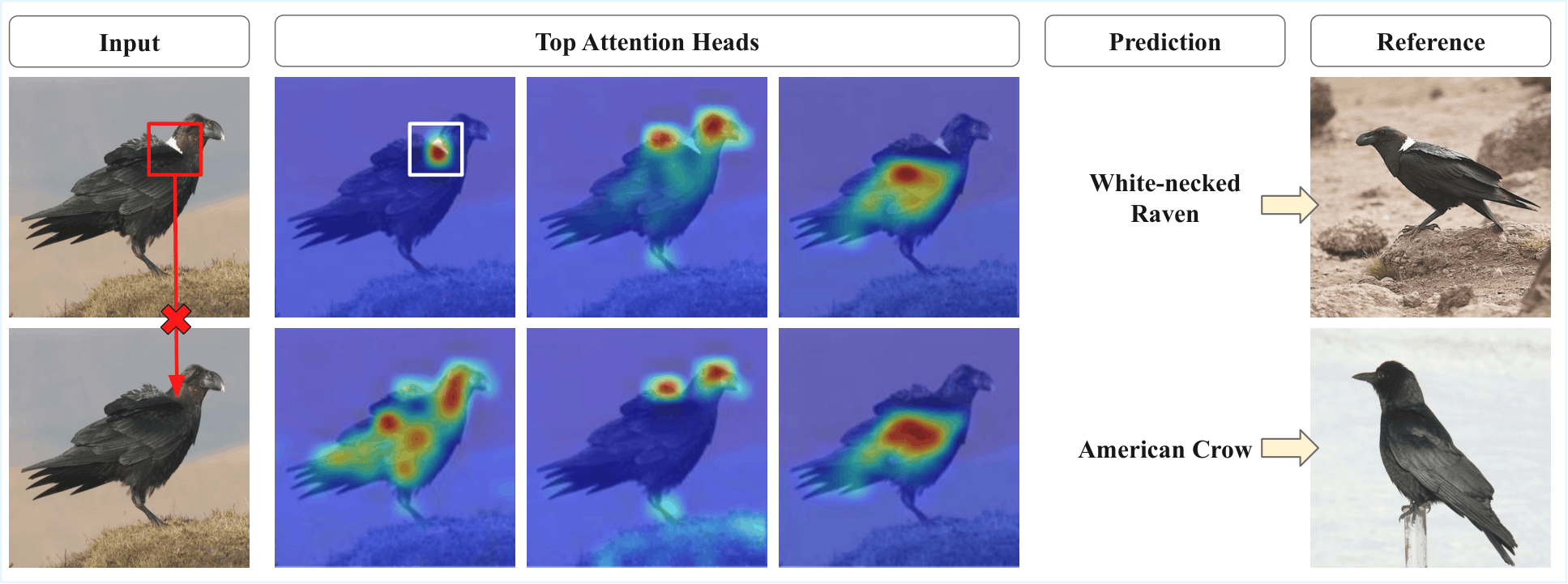}
   \caption{\textbf{Trait  manipulation.} Top row: the \textit{white-neck }trait highlighted as important for the White-necked Raven. Bottom row: after deleting this trait, the model predicts American Crow and no longer considers the white neck as important.}
   \label{trait_manupulation}
\end{figure}

\textbf{\Ours for trait manipulation.}
We validate the importance of the highlighted traits through an attribute manipulation analysis. Specifically, we select an input image of a White-necked Raven from the CUB dataset as in~\autoref{trait_manupulation}. During prediction, the model highlights the white neck as an important trait for this species. Next, we manually edit the image to remove this trait following~\cite{Cornell, Birds}, as shown in the second row. The model then predicts American Crow, which indeed resembles the modified image. This demonstrates that \Ours actually relies on such traits for species identification. We perform a similar attribute manipulation for medical tasks.  More in the Suppl. Sec.~\ref{manipulation}.


\begin{figure*}[!htbp]
  \centering
   \includegraphics[width=0.99\linewidth]{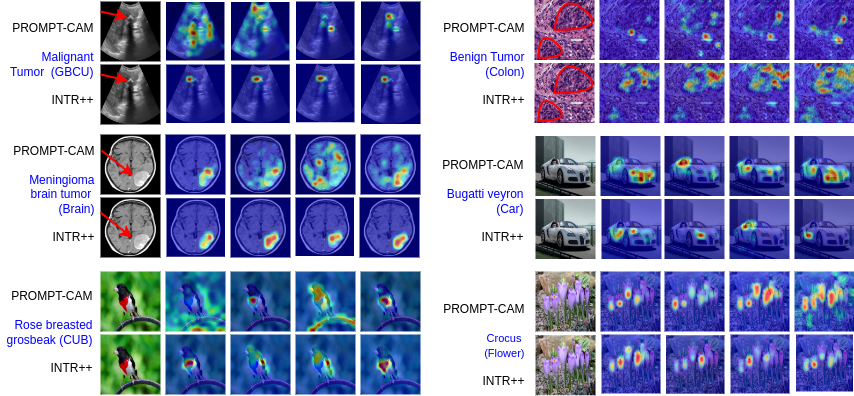}
   \vskip -2mm
   \caption{\textbf{\Ours and \OursP interpretation comparisons across all domains.} Red markings denote regions of interest.}
   \label{fig_Prompt_INTRAI}
\end{figure*}

\textbf{\Ours vs. Prompt-CAM.}
\Ours builds upon \OursP to enhance interpretability and extend applicability across domains, including specialized input domains. ~\autoref{fig_Prompt_INTRAI} illustrates a comparison between \Ours and \OursP. \Ours demonstrates substantial improvements in the medical imaging, microscopic imaging performing consistently across diverse imaging modalities, MRI, WSI, and ultrasound. For biodiversity and other tasks, both models perform comparably, although \Ours offers slightly clearer interpretations. Furthermore, \Ours yields correct predictions and interpretations in cases where \OursP fails, underscoring its  cross-domain effectiveness. More in the Suppl. Sec.~\ref{across_domain}, \ref{fail}  and \ref{additional}.

\vspace{3mm}

\begin{figure}[!htbp]
  \centering
   \includegraphics[width=0.88\linewidth]{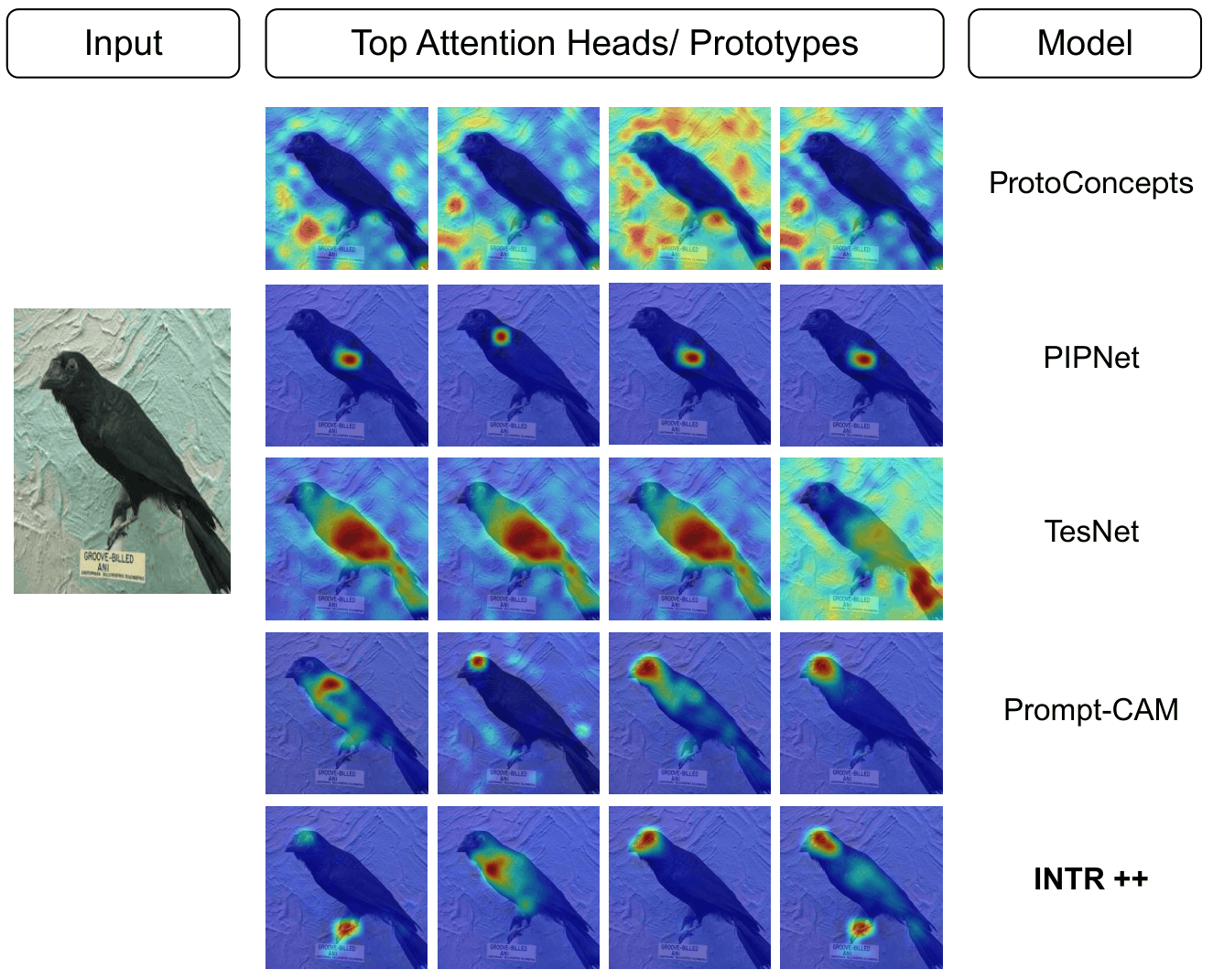}
   \vspace{-2mm}
   \caption{\textbf{Comparison of interpretable models.} Visual demonstration (heatmaps) of the four strongest
responses  on a “Groove Billed Ani” example image.}
   \label{comparision_interpretable_cub}
\end{figure}

\textbf{Qualitative comparisons of \Ours and other interpretable methods.}
We compare \Ours with various interpretable methods, including  ProtoConcepts, PIPNet TesNet and  \OursP utilizing a common backbone DINO. For simplicity, we consider a single image from the CUB dataset. \autoref{comparision_interpretable_cub}, illustrates the traits identified by each method. ProtoConcepts, PIPNet and TesNet tend to highlight only a few traits, often focusing on a single part. In contrast, \Ours produces sharper, more localized interpretations that capture greater number of discriminative traits. While \Ours and \OursP identify a comparable set of traits, \Ours offers clearer  interpretations showing its effectiveness.


\vspace{-2mm}

\begin{figure}[!htbp]
  \centering
   \includegraphics[width=0.90\linewidth]{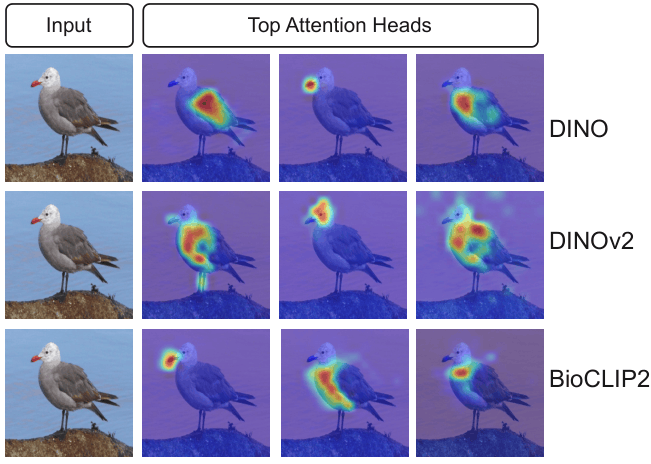}
   \caption{\textbf{\Ours on different backbones.} \Ours can identify important traits for differentiating Heermann Gull species irrespective of the backbone used.}
   \vspace{-2mm}
   \label{backbones_all}
\end{figure}

\textbf{\Ours on different backbones.}
We analyze images from the CUB dataset and visualize interpretations across different backbones, including DINO, DINOv2, and BioCLIP2. As shown in~\autoref{backbones_all}, \Ours consistently identifies distinct traits for prediction, irrespective of the backbone. \autoref{tab_accuracy_backbone} presents the accuracy comparison on dataset from different domains. More in the Suppl. Sec.~\ref{across_backbone}.



\begin{table}[!htbp]
\centering
\setlength{\tabcolsep}{3pt}

\resizebox{\linewidth}{!}{
    \begin{tabular}{llcccccc}
    \toprule
    \multicolumn{2}{l}{} & \multicolumn{1}{l}{CUB} & \multicolumn{1}{l}{Car} & 
    \multicolumn{1}{l}{Flower} & \multicolumn{1}{l}{GBCU} & 
    \multicolumn{1}{l}{Colon} & \multicolumn{1}{l}{AID} \\
    \midrule
    \multirow{3}{*}{Ours} & DINO      & 79.0 & 88.1 & 87.2 & 81.2 & 84.4 & 97.7 \\
                          & DINOv2    & 79.6 & 91.3 & 87.4 & 82.0 & 84.7 & 96.6 \\
                          & BioCLIP-2 & 91.1 & 94.2 & 96.6 & 82.3 & 84.9 & 97.4 \\
    \bottomrule
    \end{tabular}
}
\caption{\textbf{Accuracy of \Ours on different backbones.} We consider one dataset from each of the animal, object, plant and medical domain to demonstrate its effectiveness across backbones.}
\label{tab_accuracy_backbone}
\end{table}



\begin{figure}[!t]
  \centering
   \includegraphics[width=0.86\linewidth]{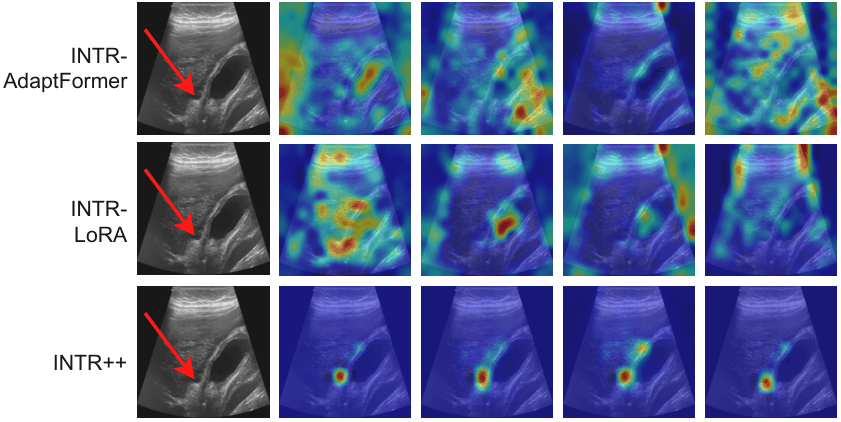}
   \vskip -2mm
   \caption{\textbf{Comparison with  adapters.} \Ours highlights important traits, while others produce noisier outputs.}
   \label{comparision_adapters1}
\end{figure}

\textbf{Comparison of \Ours with other adapter modules.} We compare \Ours with two of its variants, INTR-LoRA and INTR-AdaptFormer, which are designed using different adaptive module strategies. INTR-LoRA incorporates a LoRA-inspired adaptation module, while INTR-AdaptFormer builds upon the AdaptFormer approach. We evaluate their performance in terms of both classification accuracy and interpretability. \autoref{adapter_accuracy} presents results on one dataset from  each domains.  Moreover, we provide interpretability comparisons considering medical imaging GBCU dataset in \autoref{comparision_adapters1}. Clearly, \Ours dominates in terms of accuracy as well as interpretability. More in the Suppl. Sec.~\ref{adapters}. In addition, we validate the effectiveness of each PAM component in \Ours through the ablation study in~\autoref{ablation}. More in the Suppl. Sec.~\ref{ablation_sec}. We also provide statistical test in the Suppl. Sec.~\ref{stat}.

\begin{table}[!htbp]
\small
\centering
\setlength{\tabcolsep}{3pt}
\begin{tabular}{lcccccc}
\toprule
 & CUB & Car & Flower & GBCU & Colon & AID \\
\midrule    
INTR-MLP        & 74.0 & 67.2 & 62.6 & 51.7 & 72.4 & 89.0 \\
INTR-AdaptFormer & 72.6 & 81.0 & 76.7 & 41.0 & 80.9 & 96.8 \\  
INTR-LoRA       & 75.0 & 83.0 & 85.1 & 55.7 & 81.2 & 97.3 \\          
\Ours           & \textbf{79.0} & \textbf{88.1} & \textbf{87.2} & 
\textbf{81.2} & \textbf{84.4} & \textbf{97.7} \\           
\bottomrule  
\end{tabular}
\vskip -2mm
\caption{\textbf{Accuracy comparison.} \Ours with other adapters.}
\vskip -10pt
\label{adapter_accuracy}
\end{table}

\begin{table}[!htbp]
\small
\centering
\setlength{\tabcolsep}{2pt}
\begin{tabular}{lcccccc}
\toprule
Method & CUB & Car & Flower & GBCU & Colon & AID \\
\midrule
\Ours\ $\setminus$\ \{PCA\} & 77.3 & 87.1 & 85.1 & 80.8 & 82.5 & 97.1\\
\Ours\ $\setminus$\ \{PSA\} & 73.3 & 85.8 & 86.7 & 79.2 & 82.1 & 96.0\\
\Ours\ $\setminus$\ \{PSA, PCA\} & 72.6 & 82.8 & 83.4 & 41.0 & 74.5 & 96.9\\
\Ours & \textbf{79.0} & \textbf{88.1} & \textbf{87.2} & \textbf{81.2} & \textbf{84.4} & \textbf{97.7}\\
\bottomrule
\end{tabular}
\vskip -2mm
\caption{\textbf{Ablation study of \Ours.} The study shows the significance of each of the components of the Prompt Adapter.}
\vskip-12pt
\label{ablation}
\end{table}

\section{Conclusion}

In this paper, we proposed an interpretable transformer method, \Ours, which introduces a lightweight adaptive component, Prompt Adapter that enriches the prompts, to jointly improve prediction accuracy, interpretability, and model adaptability.  \Ours effectively resolves the trade-off between accuracy and interpretability. 
Moreover, it demonstrates strong adaptability across domains, performing consistently well on both natural and specialized image domains, thereby highlighting its generalizability.

\section*{Acknowledgments}
Our research is supported by JST Grant Number JP-MJPF2017
and International Education and Research Laboratory Program of the University of Tsukuba.
{
    \small
    \bibliographystyle{ieeenat_fullname}
    \bibliography{main}
}

\clearpage
\section*{\Huge\bfseries Supplementary}
\appendix
\vspace{5mm}
\section{Additional Details on \Ours}

\subsection{Adapter Modules}
\label{adaption_Suppl}
\begin{figure}[!htbp]
  \centering
   \includegraphics[width=0.99\linewidth]{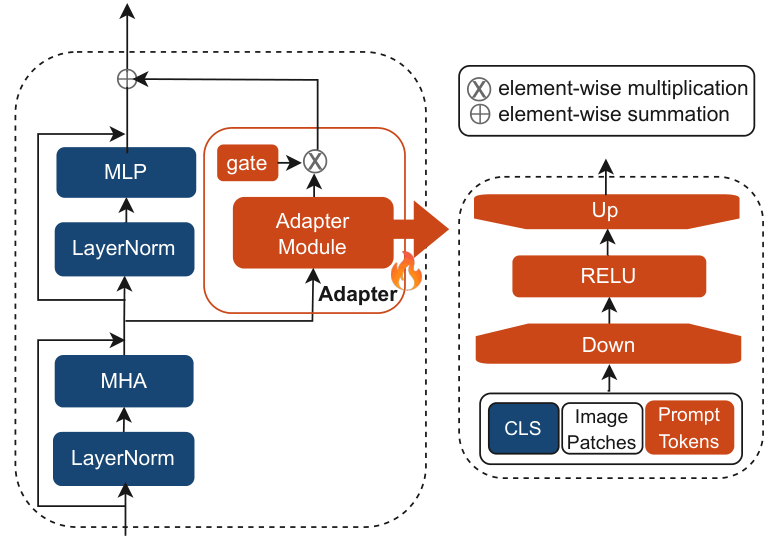}
   \caption{\textbf{Design of adapter AdaptFormer.} The module applies a lower-dimensional projection, a non-linear activation, and a projection back to the original dimension.}
   \label{flowchart_AdaptFormer}
   \vspace{-3mm}
\end{figure}

In this section, we provide the architectural details of the adapter module AdaptFormer~\cite{chen2022adaptformer}. Please see~\autoref{flowchart_AdaptFormer}. This module operates by considering all input tokens without giving any special treatment to the prompt tokens. It processes image patches and prompt tokens equally. In contrast, \Ours passes the prompt tokens through attention mechanisms, enabling them to learn domain-specific knowledge via cross-attention with the image patches. We set the projection dimension of the bottleneck layer to $64$ for all the approaches, including \Ours, AdaptFormer~\cite{chen2022adaptformer}, and LoRA~\cite{hu2022lora}, as this setting yields the best performance.


\subsection{Dataset}
\label{dataset_suppl}
We comprehensively evaluate the performance on datasets from both natural and specialized image domains. The natural image datasets include the \textbf{Animal} domain: CUB-200-2011 (\textit{CUB})~\cite{wah2011caltech} and Stanford Dogs (\textit{Dog})~\cite{khosla2011novel}; the \textbf{Plant} domain: Oxford Flowers (\textit{Flower})~\cite{nilsback2008automated} and Medicinal Leaf (\textit{MedLeaf})~\cite{roopashree2020medicinal}; and the \textbf{Object-centric} domain: Stanford Cars (\textit{Car})~\cite{krause20133d} and FGVC Aircraft (\textit{Craft})~\cite{maji2013fine}. The specialized image datasets include the \textbf{Medical imaging} domain: MRI Brain Tumors (\textit{Brain})~\cite{nickparvar2024} and Gallbladder Cancer Ultrasound (\textit{GBCU})~\cite{basu2022surpassing}; the \textbf{Microscopic imaging} domain: Colon Cancer Whole Slide Imaging (\textit{Colon})~\cite{zhuhard}; and the \textbf{Remote Sensing imagery} domain: Aerial Image Dataset (\textit{AID})~\cite{xia2017aid}. All datasets used in our experiments are publicly available. We provide the statistics of the natural and specialized image datasets from different domains in~\autoref{tab:dataset_stats_animals} and~\autoref{tab:dataset_stats_medical}, respectively.


\begin{table}[!htbp]
\centering
\resizebox{\linewidth}{!}{
    \begin{tabular}{l c c c c c c}
    \toprule
    Dataset & CUB & Dog & MedLeaf & Flower & Car & Craft \\
    \midrule
    \# Train Images & 5,994 & 12,000 & 1,455 & 2,040 & 8144 & 3334  \\
    \# Test Images  & 5,795 & 8,580  & 380   & 6,149 & 8041 & 3333 \\
    \# Classes       & 200   & 120    & 30    & 102  & 196 & 100 \\
    \bottomrule
    \end{tabular}
}
\vspace{-3mm}
\caption{\textbf{Characteristics of the natural image datasets} for the Animal, Plant, and Object-centric domains.}
\label{tab:dataset_stats_animals}
\end{table}



\begin{table}[!htbp]
\centering
\small
\resizebox{0.85\linewidth}{!}{
\begin{tabular}{lcccc}
\toprule
Dataset & GBCU & Brain & Colon & AID \\
\midrule
\# Train Images & 1133 & 5712 & 4928 & 8000 \\
\# Test Images  & 122  & 1311 & 1232 & 2000 \\
\# Classes       & 3    & 4    & 8    & 30 \\
\bottomrule
\end{tabular}
}
\vspace{-3mm}
\caption{\textbf{Characteristics of the specialized image datasets} for the Medical, Microscopic, and Remote-Sensing Imagery domains.}

\label{tab:dataset_stats_medical}
\end{table}

\subsection{Model}
\label{model_suppl}
We mostly followed the implementation details of \OursP~\cite{chowdhury2025prompt}. For all medical and microscopic imaging datasets, along with MedLeaf, training was conducted for 50 epochs, whereas the remaining datasets were trained for 120 epochs. We used the Stochastic Gradient Descent (SGD) optimizer with a momentum of $0.9$. A weight decay of $0.0$ was used for all datasets. For the medical and microscopic imaging datasets, we used a learning rate of $0.03$ with a batch size of $128$, whereas for the remaining datasets, we used a learning rate of $0.01$ with a batch size of $16$.

\section{Additional Analysis on Interpretation}

\subsection{Post-hoc Visual Comparison of \Ours with Other Methods}
\label{posthoc_suppl}
\begin{figure}[!htbp]
  \centering
   \includegraphics[width=0.9\linewidth]{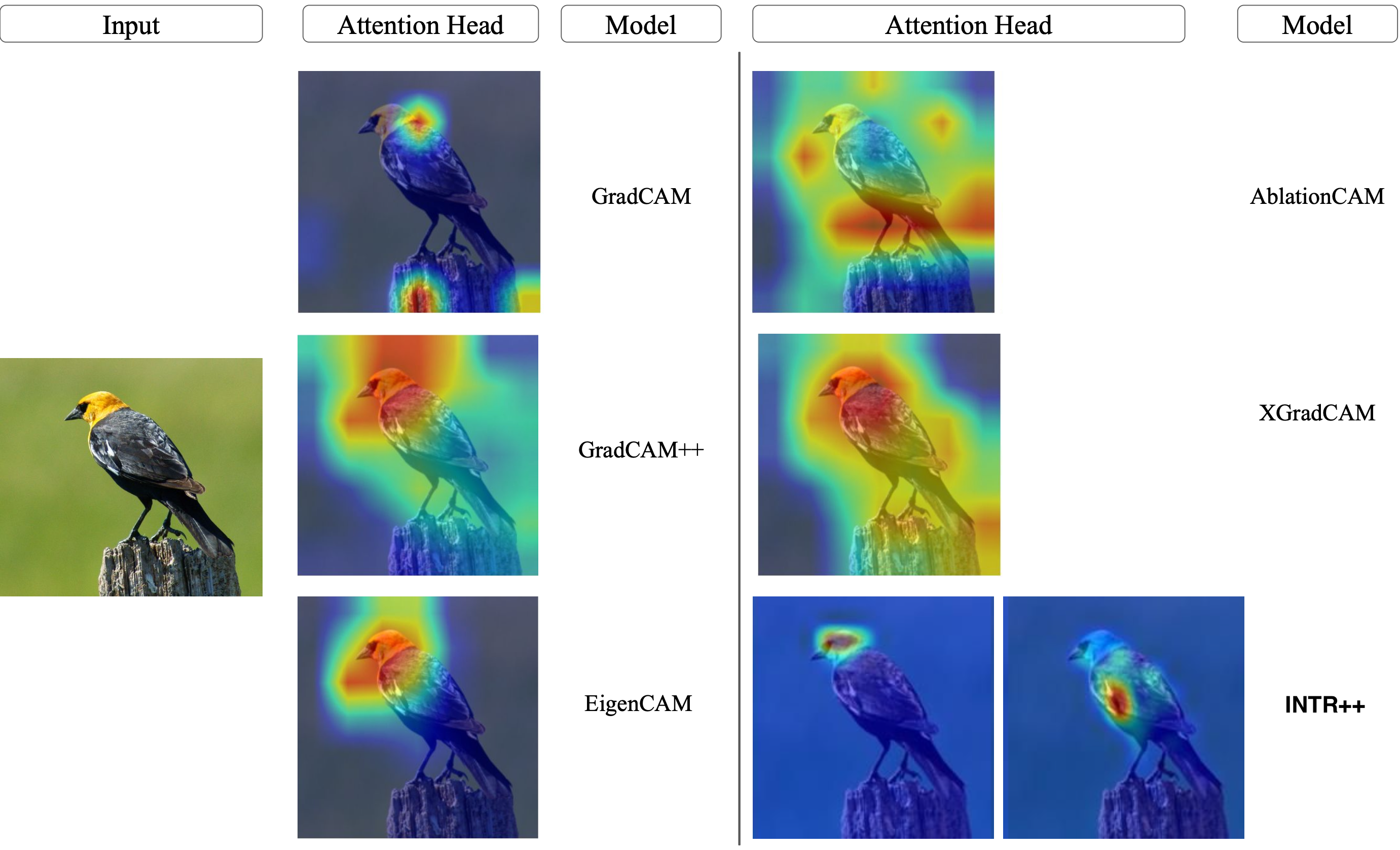}
   \caption{\textbf{Comparison of \Ours with post-hoc methods.} \Ours shows the top two attention maps for correct prediction.}
   \label{comparision_posthoc_cub}
\end{figure}

We compare \Ours with post-hoc methods using images from the CUB dataset to assess visual interpretability as shown in~\autoref{comparision_posthoc_cub}. \Ours effectively identifies the key traits necessary for correct prediction. In contrast, post-hoc methods typically highlight only one or two traits and tend to rely on global features for discrimination.
\begin{figure*}[!t]
  \centering
   \includegraphics[width=1.0\linewidth]{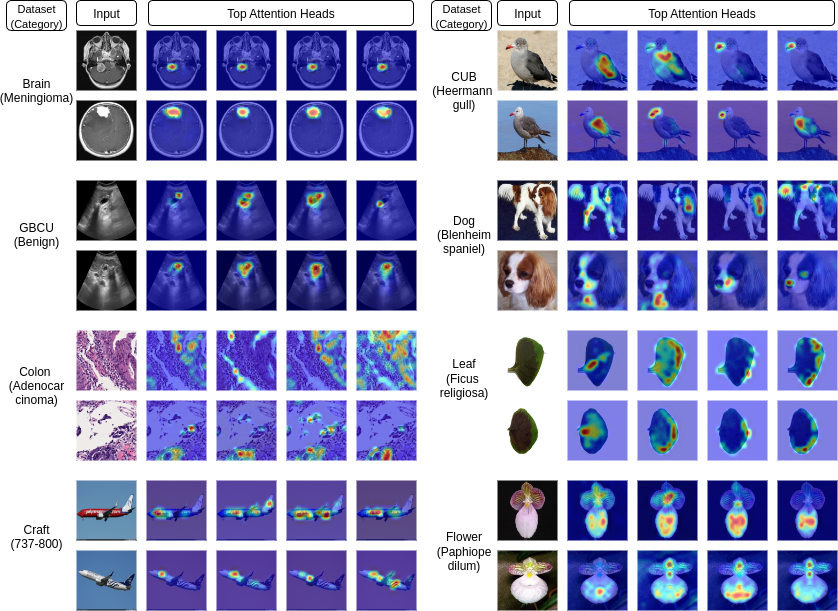}
   \vskip -2mm
   \caption{\textbf{Interpretation results of \Ours across different domains.} We show the top four attention maps for each input image. It highlights the model’s reliability in focusing on important traits or attributes for prediction across natural and specialized image domains.}
   \label{AI_CAM_dom}
\end{figure*}

\subsection{Performance of \Ours Across Domains}
\label{across_domain}
We provide the interpretation results of \Ours in~\autoref{AI_CAM_dom}. We consider the Brain and GBCU datasets from the medical imaging domain; the Colon dataset from the microscopic imaging domain; the CUB and Dog datasets from the animal domain; the MedLeaf and Flower datasets from the plant domain; and the Craft dataset from the object-centric domain for interpretation visualization and analysis. As highlighted in~\autoref{AI_CAM_dom}, for the medical and microscopic imaging domains, the model consistently identifies important area of interest. For the animal and plant domains, \Ours consistently highlights the discriminative traits necessary to distinguish among species. Moreover, for the object-centric domain, the model highlights the important regions for its prediction. These results demonstrate the generalization of \Ours across both natural and specialized image domains. They further highlight its adaptability and applicability across diverse domains compared to existing interpretable methods.

\subsection{Comparison of \Ours with Interpretable Methods for Specialized Images}
\label{com_Sp_interp}

\begin{figure}[!htbp]
  \centering
   \includegraphics[width=0.9\linewidth]{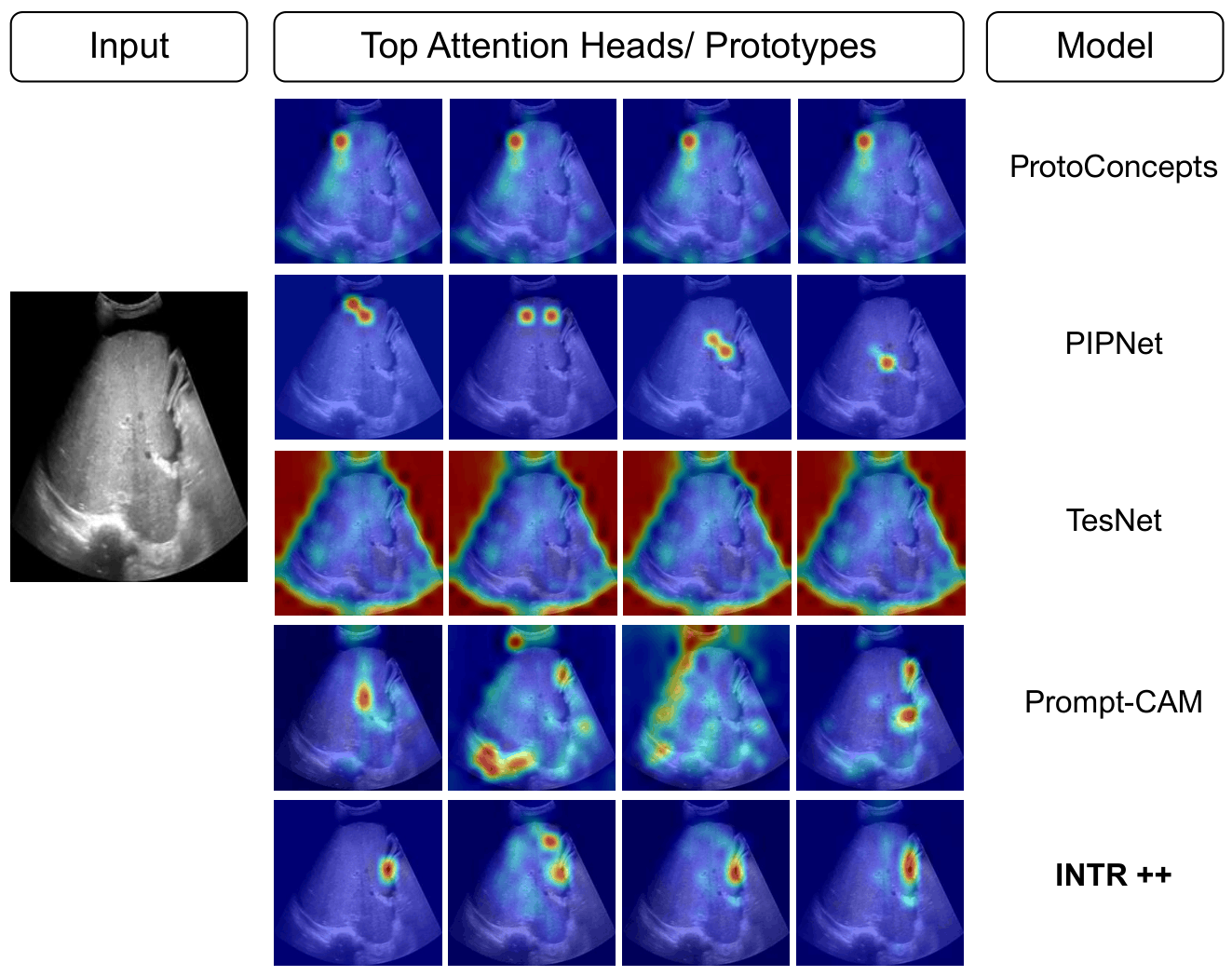}
   \caption{\textbf{Interpretation comparisons on specialized images.} We compare \Ours with other interpretable methods on the GBCU dataset for the \textit{benign} category.}
   \label{sp_inter_gbcu}
\end{figure}

We analyze the interpretability of specialized image inputs using GBCU as a representative dataset. We compare \Ours with interpretable methods, ProtoConcepts, PIPNet, TesNet and \OursP. \autoref{sp_inter_gbcu} demonstrates the superiority of \Ours, for specialized image inputs, as it clearly identifies the relevant regions of interest, whereas the other methods focus on less relevant regions.

\subsection{Comparison of \Ours and \OursP on Natural Image Tasks}
\label{fail}

\begin{figure*}[!htbp]
  \centering
   \includegraphics[width=0.99\linewidth]{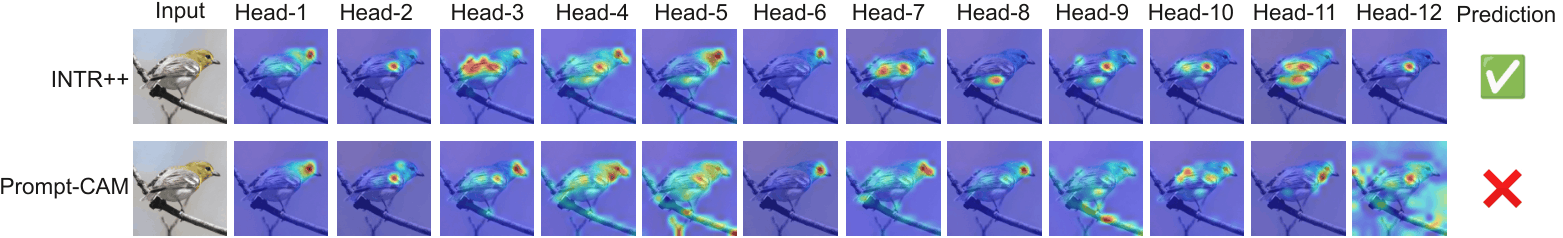}
\vspace{-3mm}
   \caption{\textbf{\Ours predicts correctly where \OursP fails.} We visualize all twelve attention heads for both approaches. \Ours effectively captures the key traits and yields accurate predictions, whereas \OursP predominantly attends to redundant traits across multiple heads and includes noisy regions, resulting in incorrect predictions.}
   \label{prompt_cam_fails}
\end{figure*}

On natural image tasks, \Ours and \OursP often achieve comparable levels of interpretability. However, \Ours consistently exhibits superior predictive performance. As illustrated in~\autoref{prompt_cam_fails}, \Ours produces accurate predictions while providing reliable interpretations, even in challenging cases where \OursP fails, demonstrating the overall effectiveness of \Ours.

\subsection{Additional Comparison  between \Ours and \textbf{\OursP}}
\label{additional}
\begin{figure}[!htbp]
  \centering
   \includegraphics[width=0.90\linewidth]{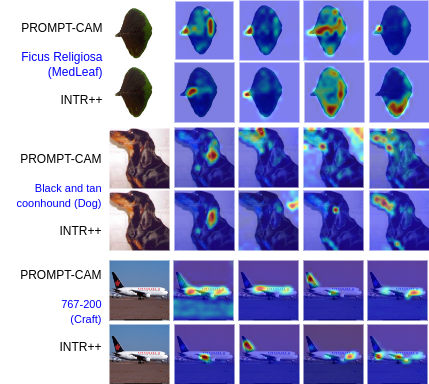}
   \vskip -2mm
   \caption{\textbf{Interpretation of \Ours and \OursP.} We show comparisons on the MedLeaf, Dog and Craft datasets. 
   }
   \vspace{-3mm}
   \label{AI_CAM_rem}
\end{figure}

We provide additional comparisons between \Ours and \OursP on the MedLeaf, Dog and Craft datasets. Please refer to~\autoref{AI_CAM_rem}. These datasets belong to the natural image domain. Therefore, the interpretation results of both methods are comparable. However, the overall adaptability and performance of \Ours make it better suited for interpretable image classification.

\subsection{Attribute Manipulation}
\label{manipulation}
\begin{figure}[!htbp]
  \centering
   \includegraphics[width=0.99\linewidth]{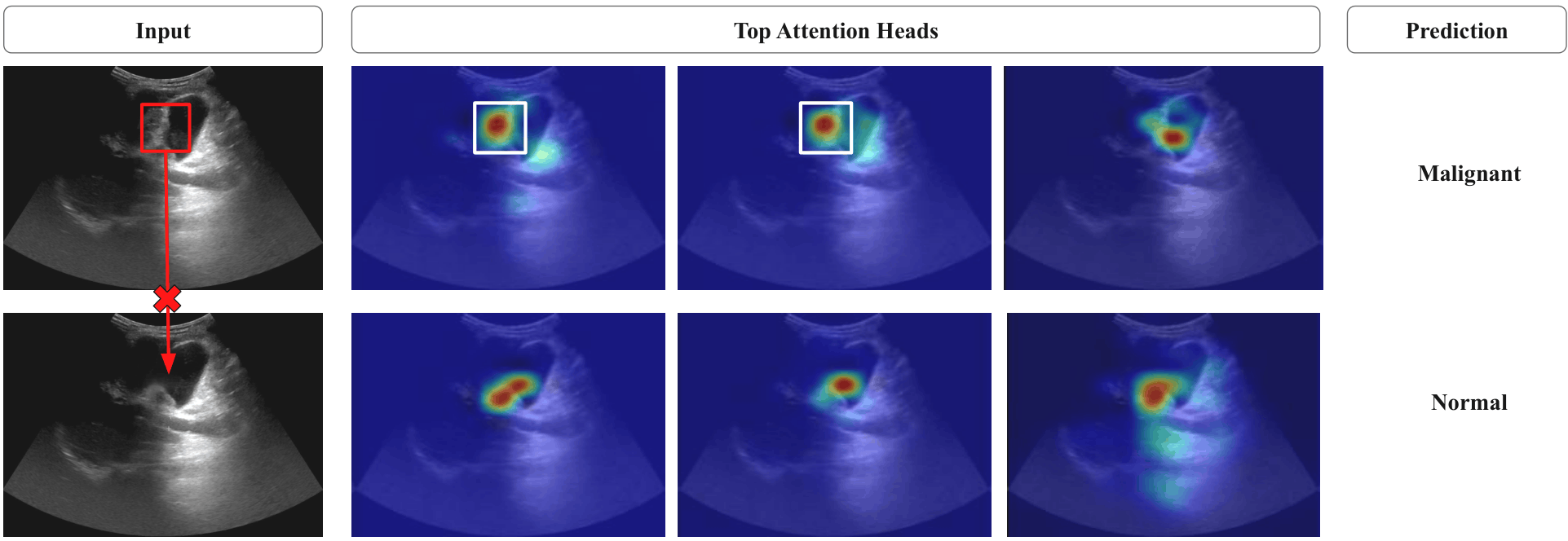}
    \vspace{-3mm}
   \caption{\textbf{Attribute manipulation.} The top row shows the important cancerous region for a malignant case; deleting it leads to a normal prediction, with attention shifting elsewhere.}
   \label{trait_manupulation_GBCU}
\end{figure}

We validate the importance of the highlighted regions in medical imaging through attribute manipulation analysis. Specifically, we select a malignant tumor image from the GBCU dataset and identify its critical region using \Ours, as shown in~\autoref{trait_manupulation_GBCU}. We then manually edit the image to remove the identified cancerous region and re-evaluate it using the model. The modified image is subsequently classified as normal, demonstrating the precision and significance of the region identified by \Ours. 

Additionally, we quantitatively analyze this study by considering a few species from the CUB dataset. We randomly selected 10 images each from the Red-winged Blackbird, Baltimore Oriole, and White-necked Raven classes and manually manipulated them to resemble American Crow, Orchard Oriole, and American Crow, respectively. The trait manipulation was performed following the guidelines from Cornell Lab\footnote{https://www.birds.cornell.edu/home} and Birds of the World\footnote{https://birdsoftheworld.org/bow/home}. We passed both the original and manipulated images through \Ours, \OursP, and PIP-Net and observed that the models correctly identified the manipulated traits with accuracies of 97.7\%, 97.7\%, and 90.0\%, respectively. We consider a manipulated trait to be correctly detected if the model predicts the target category corresponding to the manipulated image.
 


\section{ Additional Analysis  on Performance}
\subsection{Performance Comparisons}
\label{accu_suppl}
We compared \Ours with existing interpretable methods, including ProtoConcepts, PIPNet, and TesNet, based on their predictive performance measured by classification accuracy. \autoref{acc_interpretable} demonstrates the consistent superiority of \Ours over existing methods across diverse domains. All the results reported here are obtained using the DINO backbone.

\begin{table}[!htbp]
\small
\centering
\setlength{\tabcolsep}{4pt}
\begin{tabular}{lcccccc}
\toprule
 & CUB & Car & Flower & GBCU & Colon & AID \\
\midrule

ProtoConcepts & 68.3 & 78.7 & 85.5 & 41.8 & 78.9 & 95.1 \\
PIPNet   & 78.8   & 84.4 & \textbf{88.2} & 79.9 & 81.8 & 95.4 \\
TesNet      & 58.9  & 58.5 & 50.3 & 77.8 & 80.2 & 94.9 \\
\Ours     & \textbf{79.0}    & \textbf{88.1} & 87.2 & \textbf{81.2} & \textbf{84.4} & \textbf{97.7} \\
\bottomrule
\end{tabular}
\vskip -2mm
\caption{\textbf{Accuracy comparisons.} \Ours with other interpretable models. We consider one dataset from each domain.}
\vskip -10pt
\label{acc_interpretable}
\end{table}

\subsection{Interpretability Comparisons}
\label{Interpretability_Comparisons}
We compare \Ours with existing interpretable methods, including ProtoConcepts, PIPNet, and TesNet, in terms of interpretability using the insertion and deletion metrics. \autoref{tab_insertion_deletion_1} and \autoref{tab_insertion_deletion_2} show the results for natural and specialized image input datasets, respectively. The results clearly demonstrate the superior interpretability of \Ours over other interpretable methods.

\begin{table}[!htbp]
\vskip-5pt
\centering
\footnotesize
\setlength{\tabcolsep}{3pt}
\begin{tabular}{lcccccc}
\toprule
\multirow{2}{*}{Method} 
& \multicolumn{2}{c}{CUB} 
& \multicolumn{2}{c}{Flower} 
& \multicolumn{2}{c}{Car} \\
& Insert.$\uparrow$ & Delet.$\downarrow$
& Insert.$\uparrow$ & Delet.$\downarrow$
& Insert.$\uparrow$ & Delet.$\downarrow$ \\
\midrule

ProtoConcepts   & 0.11 & 0.76 & 0.36 & 0.61 & 0.48 & 0.15 \\
PIPNet          & 0.59 & 0.10 & \textbf{0.76} & 0.29 & 0.79 & 0.12 \\
TesNet          & 0.43 & 0.18 & 0.24 & \textbf{0.16} & 0.50 & 0.09 \\
\textbf{\Ours}  & \textbf{0.62} & \textbf{0.08} 
                & 0.66 & 0.18
                & \textbf{0.82} & \textbf{0.04} \\
\bottomrule
\end{tabular}
\vspace{-3mm}
\caption{\small \textbf{ Evaluation of interpretation for natural image datasets.} Comparisons  among interpretable methods }
\label{tab_insertion_deletion_1}
\end{table}

\begin{table}[!htbp]
\vskip-5pt
\centering
\footnotesize
\setlength{\tabcolsep}{3pt}
\begin{tabular}{lcccccc}
\toprule
\multirow{2}{*}{Method} 
& \multicolumn{2}{c}{GBCU} 
& \multicolumn{2}{c}{Colon} 
& \multicolumn{2}{c}{AID} \\
& Insert.$\uparrow$ & Delet.$\downarrow$
& Insert.$\uparrow$ & Delet.$\downarrow$
& Insert.$\uparrow$ & Delet.$\downarrow$ \\
\midrule

ProtoConcepts   & 0.52 & 0.45 & 0.45 & 0.55 & 0.50 & 0.66 \\
PIPNet          & 0.63 & 0.52 & 0.75 & 0.59 & \textbf{0.66} & 0.64 \\
TesNet          & 0.58 & 0.55 & 0.75 & 0.43 & 0.65 & 0.66 \\
\textbf{\Ours}  & \textbf{0.65} & \textbf{0.19} 
                & \textbf{0.77} & \textbf{0.39}
                & \textbf{0.66} & \textbf{0.50} \\
\bottomrule
\end{tabular}
\vspace{-3mm}
\caption{\small \textbf{ Evaluation of interpretation for specialized image datasets.} Comparisons  among interpretable methods }
\label{tab_insertion_deletion_2}
\end{table}

\subsection{Interpretation of \Ours Across Backbones}
\label{across_backbone}
A key strength of \Ours is its robustness across different backbone models. In this work, we primarily use the DINO backbone for analysis. However, we also evaluate \Ours with other backbones, including DINOV2~\cite{oquab2023dinov2} and BioCLIP2~\cite{gu2025bioclip}. \autoref{tab_accuracy_backbone} reports the classification accuracy on the CUB, Car, Flower, GBCU, Colon and AID datasets for each backbone. The observed performance improvements are primarily attributed to the backbone models being pre-trained on larger datasets, while the interpretability of \Ours remains consistent regardless of the backbone. As shown in~\autoref{backbones_all}, which presents interpretation results for the Heermann Gull species from the CUB dataset, \Ours consistently identifies distinct discriminative traits irrespective of the backbone.

\section{Additional Results}

\subsection{Ablation Study}
\label{ablation_sec}
\begin{figure}[!htbp]
    \centering
    \includegraphics[width=0.96\linewidth]{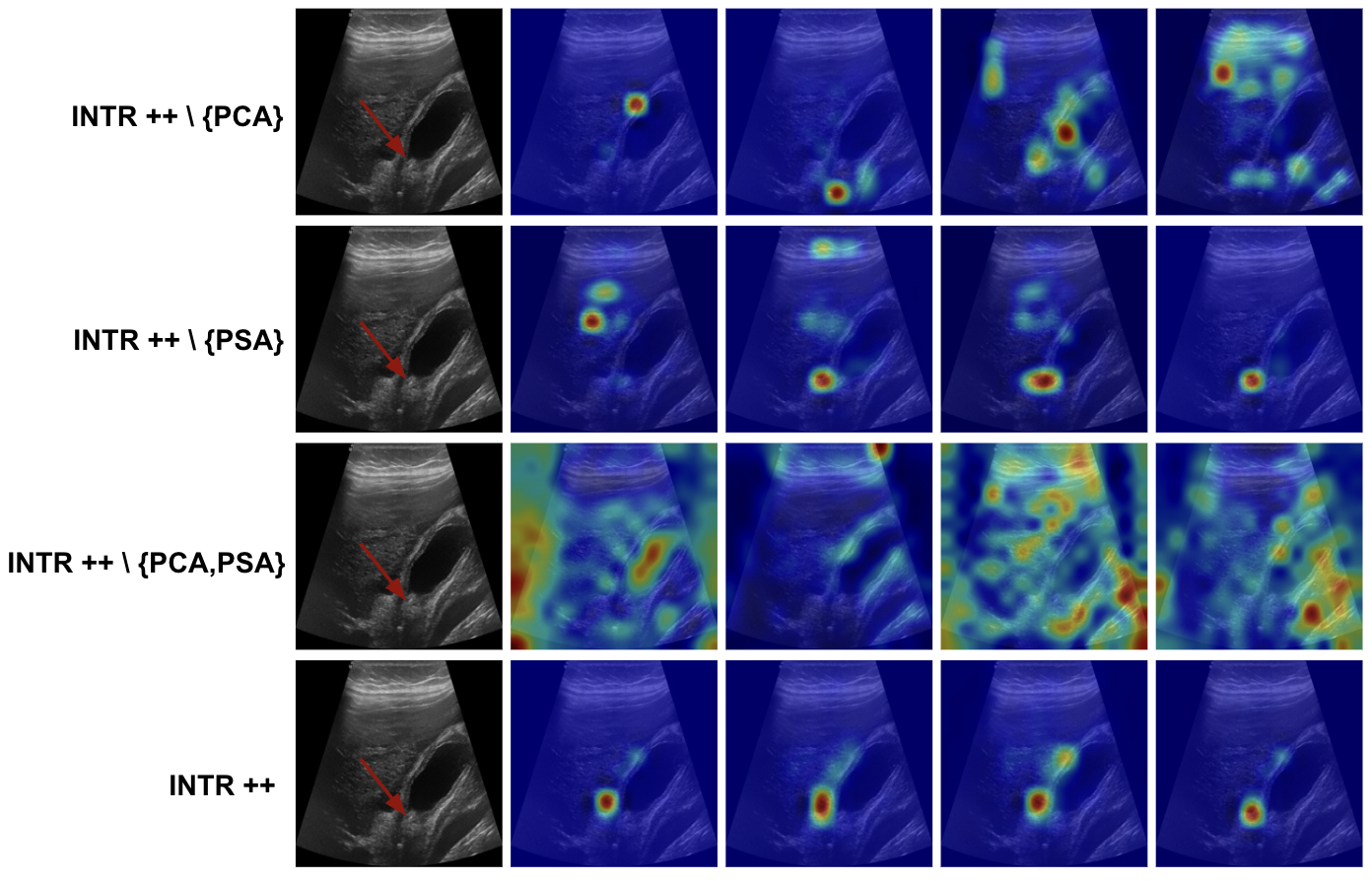}
    \vskip -3mm
    \caption{\textbf{Ablation study on GBCU dataset.} Visual comparison demonstrating the importance of each component in the PAM of \Ours on the specialized images input.}
    \label{gbcu_ablation}
\end{figure}

\begin{figure}[!htbp]
    \centering
    \includegraphics[width=0.96\linewidth]{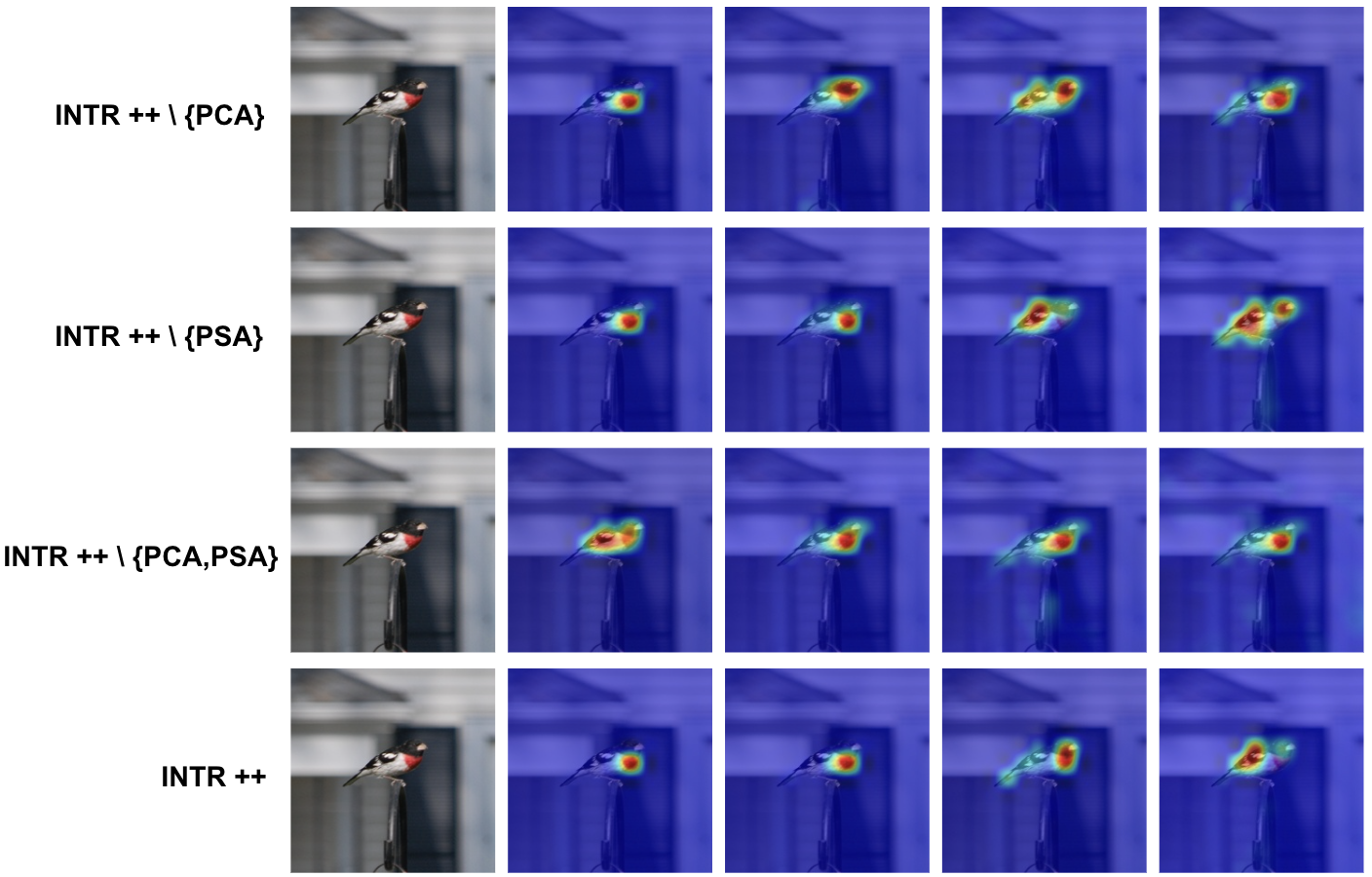}
    \vskip -3mm
    \caption{\textbf{Ablation study on CUB dataset.} Visual comparison demonstrating the importance of each component in the PAM of \Ours on the natural images input.}
    \label{cub_ablation}
\end{figure}

We illustrate the interpretation after eliminating one component at a time from the PAM module of \Ours in~\autoref{gbcu_ablation} and ~\autoref{cub_ablation}. Clearly, the results demonstrate the importance of each component, particularly the role of PCA. Overall, the complete PAM in \Ours (last row) provides the most appropriate interpretation, as it is able to capture important regions in~\autoref{gbcu_ablation} for the \textit{malignant} case in the GBCU dataset and different attributes in ~\autoref{cub_ablation} for the \textit{Rose-breasted Grosbeak} species in the CUB dataset. .

\subsection{Role of Gate g} 
\label{gate_value}

\begin{figure}[!htbp]
    \centering
    \includegraphics[width=0.96\linewidth]{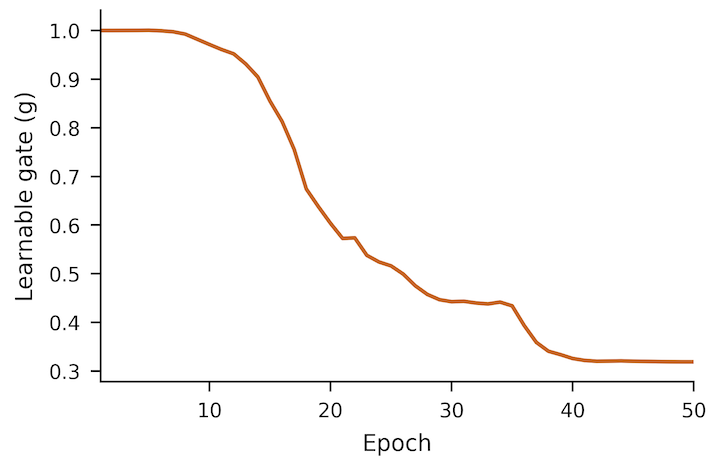}
    \vskip -3mm
    \caption{\textbf{Learned gate values over training epochs.} The plot shows the  last layer gate values of \Ours on  GBCU dataset.}
    \label{gate_gbcu}
\end{figure}

Each adapter output is integrated into the backbone through a learnable gating parameter, g, which controls the fraction of the adapter information added to the original transformer output. A gate value of 0 reduces the model to Prompt-CAM, while a gate value of 1 fully incorporates the adapter output. The gate is initialized to g=1, allowing full adapter contribution during the early stages of training to facilitate effective domain alignment. As training progresses, g is learned jointly with the adapter parameters and gradually decreases, converging to a stable value. This behavior indicates that the model initially relies on stronger adaptive corrections and subsequently attenuates the adapter contribution as the representations become aligned, resulting in controlled and stable integration without overwhelming the pretrained backbone. Please refer to~\autoref{gate} for the CUB dataset. Additionally, ~\autoref{gate_gbcu} shows the gate scores over the training epochs for the GBCU dataset.


\subsection{Projection Dimension of the Adaptive Module}
\label{projection_dimension}
\begin{figure}[!htbp]
  \centering
   \includegraphics[width=0.88\linewidth]{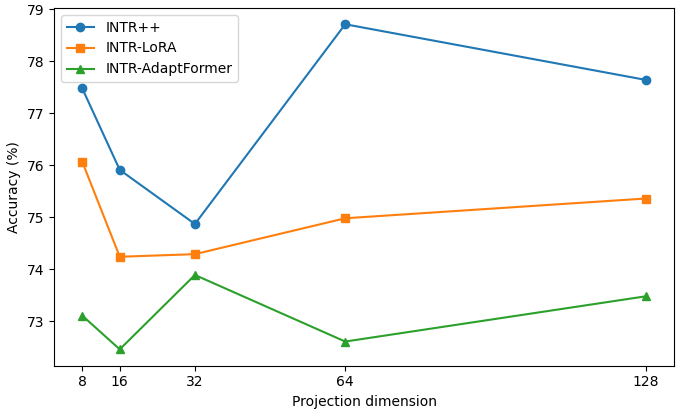}
   \caption{\textbf{Projection size comparison.} Performance comparison with different projection sizes for \Ours, INTR-LoRA, and INTR-AdaptFormer for the CUB dataset. }
   \label{bprojection_dims}
\end{figure}

We evaluate different low-dimensional projection sizes for the adaptive modules, as $8$, $16$, $32$, $64$, and $128$. As shown in~\autoref{bprojection_dims}, all methods achieve their best performance with a projection dimension of $64$. Furthermore, \Ours consistently outperforms the other methods across all projection dimensions.

\subsection{Comparisons among Different Adapters}
\label{adapters}

\begin{figure}[!htbp]
  \centering
   \includegraphics[width=1.0\linewidth]{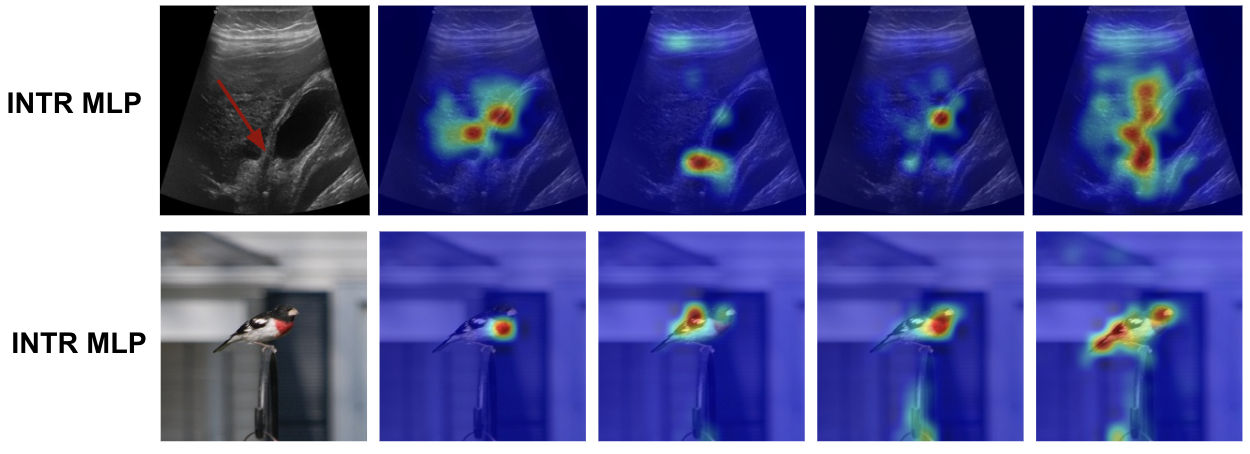}
   \vskip -3mm
   \caption{\textbf{Interpretation of MLP-based adapters.} Top attributes for an image from the CUB and GBCU datasets. }
   \label{mlp}
\end{figure}

We analyze the adaptive behavior of different adapter designs, including MLP, LoRA, AdaptFormer, and PAM (\Ours), using the cosine similarity between backbone-only and adapter-augmented representations. Across all methods, image patch tokens exhibit high similarity to the backbone representations, except for MLP, which disrupts the pretrained representations as illustrated in~\autoref{embeddings}. In contrast, prompt-wise analysis reveals that PAM induces larger representational shifts between the backbone-only and adapter-augmented representations. This behavior arises from the design of PAM, which explicitly aligns prompts with salient input characteristics through prompt-centric interactions. LoRA and AdaptFormer, on the other hand, produce prompt representations that remain closer to the original backbone representations. As a result, PAM not only adaptively adjusts the pretrained representations but also enables the prompts to better capture input characteristics, leading to improved adaptability across both natural and specialized image domains while enhancing accuracy and interpretability. \autoref{mlp} additionally illustrates the poor interpretability of MLP-based adapters, which capture noise and only a single attribute across different heads.


\section{Statistical and Computational Analysis}

\subsection{Statistical Analysis}
\label{stat}

To perform statistical testing, we consider CUB and GBCU datasets  and run each method ten times with different random seeds. We consider \Ours and \OursP and perform the Wilcoxon signed-rank test\footnote{Wilcoxon, Frank. "Individual comparisons by ranking methods." Biometrics bulletin 1.6 (1945): 80-83.}. The results reported in~\autoref{tab_stat} demonstrate the statistical significance of \Ours.

\begin{table}[!htbp]
\small
\centering
\begin{tabular}{lccc}
\toprule
Dataset & \Ours & \OursP & $p$-value \\
\midrule
CUB    & 78.2 $\pm$ 1.1 & 74.3 $\pm$ 2.21 & 0.011 \\
GBCU   & 80.4 $\pm$ 1.2 & 62.8 $\pm$ 1.71 & 0.002 \\
\bottomrule
\end{tabular}
\vskip -2mm
\caption{\textbf{Statistical analysis.} Mean $\pm$ standard deviation over ten runs. $p$-values are from the paired Wilcoxon signed-rank test between \Ours and Prompt-CAM on CUB and GBCU datasets.}
\vskip -10pt
\label{tab_stat}
\end{table}

\subsection{Computational Analysis}
\label{computation}
We compare the computational efficiency of interpretable methods, including \Ours, \OursP, ProtoConcepts, PIPNet, and TesNet, in terms of the number of learnable parameters using DINO as the backbone. As shown in~\autoref{param}, \Ours and \OursP are highly parameter-efficient. \Ours requires less than 3\% of the learnable parameters of existing methods. \OursP requires slightly less learnable parameters than \Ours; however, this marginal parameter overhead is small relative to the performance gains achieved by \Ours across natural and specialized image domains.

\begin{table}[!htbp]
\small
\centering
\setlength{\tabcolsep}{2pt}
\begin{tabular}{ccccc}
\toprule
ProtoConcepts & PIPNet & TesNet & \OursP & \Ours \\
\midrule
$\approx85$--$86$ & $\approx86$--$87$ & $\approx86$--$87$ &
$\approx0.1$--$1.0$ & $\approx1.2$--$3.4$ \\
\bottomrule
\end{tabular}
\vskip -2mm
\caption{\textbf{Number of learnable parameters (in millions).} Comparison of interpretable methods in terms of parameter efficiency using DINO as the backbone. Values indicate the approximate range across all datasets.}
\vskip -10pt
\label{param}
\end{table}


\end{document}